\documentclass[journal]{IEEEtran}
\usepackage{graphicx}
\ifCLASSINFOpdf
\else
\fi
\usepackage{amsmath}
\usepackage{algorithmic}
\usepackage{algorithm}
\usepackage{array}

\usepackage{subfig}
\usepackage{booktabs}
\usepackage{xcolor}
\usepackage{amssymb}
\usepackage{hyperref}
\usepackage{listings}

\lstdefinestyle{lfonts}{
  basicstyle   = \footnotesize\ttfamily,
  stringstyle  = \color{purple},
  keywordstyle = \color{blue!60!black}\bfseries,
  commentstyle = \color{olive}\scshape,
}
\lstdefinestyle{lnumbers}{
  numbers     = left,
  numberstyle = \tiny,
  numbersep   = 1em,
  firstnumber = 1,
  stepnumber  = 1,
}
\lstdefinestyle{llayout}{
  breaklines       = true,
  tabsize          = 2,
  columns          = flexible,
}
\lstdefinestyle{lgeometry}{
  xleftmargin      = 20pt,
  xrightmargin     = 0pt,
  frame            = tb,
  framesep         = \fboxsep,
  framexleftmargin = 20pt,
}
\lstdefinestyle{lgeneral}{
  style = lfonts,
  style = lnumbers,
  style = llayout,
  style = lgeometry,
}
\lstdefinestyle{python}{
    language = {Python},
    style    = lgeneral,
}

\begin{document}
% \title{Efficient Reinforcement Learning for StarCraft with a Hand-designed Forward Model and XfrNet}
\title{Efficient Reinforcement Learning for StarCraft by Abstract Forward Models and Transfer Learning}
% ~\IEEEmembership{Member,~IEEE,}
% ~\IEEEmembership{Member,~IEEE,}
\author{Ruo-Ze~Liu,
        Haifeng~Guo,
        Xiaozhong~Ji,
        Yang~Yu,~\IEEEmembership{Member,~IEEE,}
        Zhen-Jia~Pang,
        Zitai~Xiao,
        Yuzhou~Wu,
        Tong~Lu,~\IEEEmembership{Member,~IEEE,}% <-this % stops a space
\thanks{Y. Yu and T. Lu were corresponding authors. They were with the National Key Laboratory for Novel Software Technology, Nanjing University e-mail: yuy@nju.edu.cn, lutong@nju.edu.cn.}% <-this % stops a space
\thanks{R-Z. Liu, H. Guo, X. Ji, Z-J. Pang, Z. Xiao, Y. Wu were students with Nanjing University.}% <-this % stops a space
\thanks{Manuscript received April 19, 2020; revised December 31, 2020.}}


\markboth{Journal of \LaTeX\ Class Files,~Vol.~14, No.~8, August~2020}%
{Shell \MakeLowercase{\textit{et al.}}: Bare Demo of IEEEtran.cls for IEEE Journals}



\maketitle

\begin{abstract}
Injecting human knowledge is an effective way to accelerate reinforcement learning (RL). However, these methods are underexplored. This paper presents our discovery that an abstract forward model (Thought-game (TG)) combined with transfer learning is an effective way. We take StarCraft II as the study environment. With the help of a designed TG, the agent can learn a 99\% win-rate on a 64$\times$64 map against the Level-7 built-in AI, using only 1.08 hours in a single commercial machine. We also show that the TG method is not as restrictive as it was thought to be. It can work with roughly designed TGs, and can also be useful when the environment changes. Comparing with previous model-based RL, we show TG is more effective. We also present a TG hypothesis that gives the influence of fidelity levels of TG. For real games that have unequal state and action spaces, we proposed a novel XfrNet of which usefulness is validated while achieving a 90\% win-rate against the cheating Level-10 AI. We argue the TG method might shed light on further studies of efficient RL with human knowledge.
\end{abstract}

% , on which the TG method is more effective by Comparing with the previous model-based RL algorithms. 

%  For real games that have unequal state and action spaces, we proposed an XfrNet and validate it by experiments on SC2. We also show a TG hypothesis which give the influence of different fidelity levels of TG.

% for the TG making the policy can transfer to a real game with unequal state spaces and action spaces. A TG hypothesis is then given showing the influence of the different fidelity levels of TG.

% Finally, we validate the effectiveness of the TG method also on StarCraft I and a real-world hydropower task. 


% \fi
\IEEEpeerreviewmaketitle

\section{Introduction}
\IEEEPARstart{I}{n} recent years, the combination of reinforcement learning (RL)~\cite{sutton1998introduction} and deep learning (DL)~\cite{DeepLearning2015Nature}, the deep reinforcement learning (DRL), has received increasing attention, particularly in learning to play games~\cite{mnih2015human}. The combination of DRL and Monte-Carlo tree search has conquered the game of Go~\cite{silver2017mastering}. After that, some researchers shifted their attention to real-time strategy (RTS) games, e.g., RL has been applied to StarCraft I (SC1)~\cite{bilibiliSC} \& II (SC2)~\cite{AlphaStarNature}, and Dota2. However, huge computing resources and a long time is required by the above approaches, e.g., TStarBot~\cite{Sun2018tsbot} uses $3840$ CPU cores and OpenAIdota2 utilizes $128400$ ones. Also notice that the training time of OpenAIdota2 is calculated in months, and full training in~\cite{AlphaStarNature} takes $44$ days. Such costs make RL approaches impractical for real-world tasks.

Meanwhile, it has been widely recognized that injecting human knowledge is an effective way to improve the efficiency of RL. Behavior cloning~\cite{behavioralCloning}, one kind of imitation learning (IL), is an often adopted approach to boost the starting performance of RL~\cite{AlphaStarNature}. It should not be ignored that human knowledge has also been utilized in choosing the best policy model structure, designing the best reward function, and deciding the best learning algorithm hyper-parameters. However, the investigation of effective knowledge injection methods is an overlooked issue and desires more exploration.

This paper investigates, from the perspective of model-based RL (MB-RL), the knowledge injection method by designing an abstracted and simplified model of environment. Such models are called forward models~\cite{FB2SC1} or world models~\cite{Ha2018worldmodelNIPS} in previous literatures. Different from them, the model we investigate here is a hand-designed one, moreover, it needs not to be identical to the original one as much as possible. Hence, our model can be seen as an abstract forward model. In this paper, we give it a simple name \textit{Thought-game} (TG). This name draws on the thinking mode of human players in RTS games. After playing some games, players will build models of the game in their minds. The models can be used to pre-train their game abilities before they go to real games. Previous, it is believed that an accurately reconstructed model can be helpful to accelerate the RL. On the one hand, however, learning an accurate model can be quite tricky; on the other hand, human experts often have a bunch of experiences that can abstractly outline the environment. Therefore, it is interesting to investigate whether an abstracted and drastically simplified environment model constructed from human experience can help RL.

% The name of TG is for clearly distinguishing it from the normal forward model

\begin{figure}[h!]
    \begin{minipage}[t]{\linewidth}
        \centering
        \includegraphics[width=0.85\columnwidth]{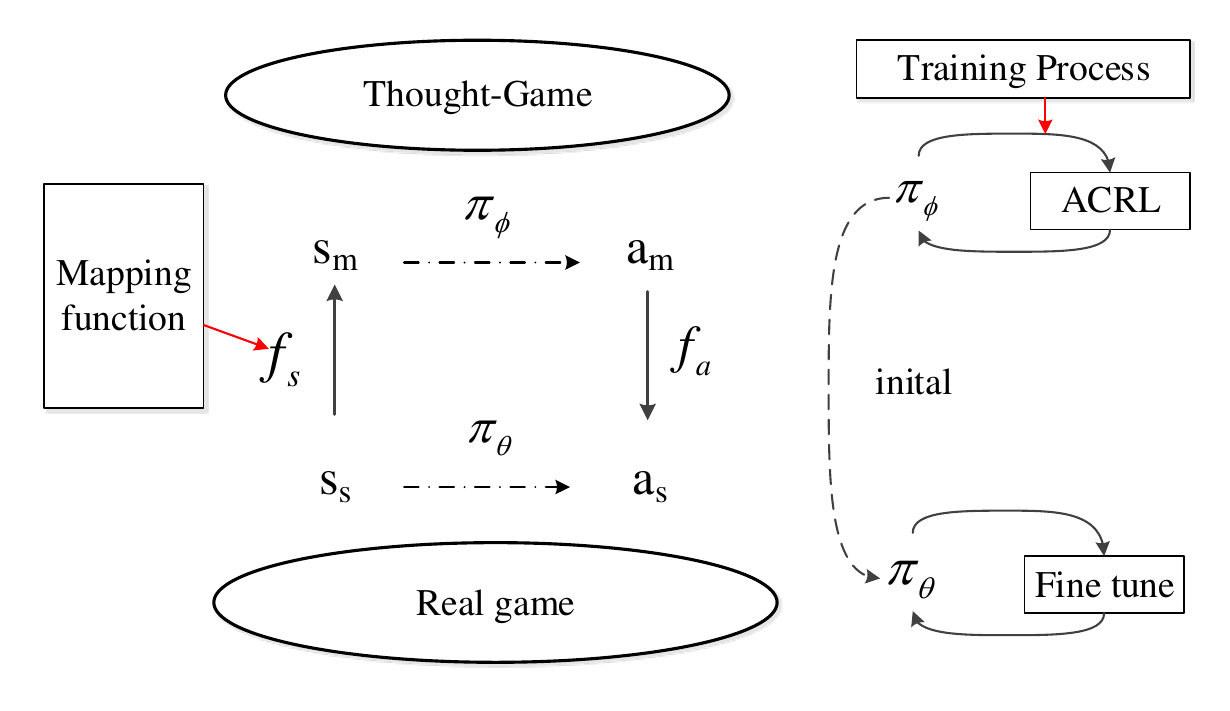}
        \caption{Thought-Game Architecture. $S_m$ and $S_s$ are the state spaces for TG and RG respectively. $A_m$ and $A_s$ are the action spaces while $\pi_{\phi}$ and $\pi_{\theta}$ are the policies. The dashed line in the right means the weight-parameters in $\pi_{\theta}$ are initialed as ones in $\pi_{\phi}$ before finetune training.}
        \label{fig:Thought-Game}
    \end{minipage}
\end{figure}

We take SC2 as the study environment, which is a challenging game for RL. We design a TG and propose a training algorithm TTG (Training with TG). As described in the right of Fig.~\ref{fig:Thought-Game}, we first train the agent using an automated curriculum reinforcement learning (ACRL) algorithm in the TG, and then using transfer learning (TL) algorithm \textit{finetune}~\cite{Finetune2014} to transfer it to the real game (RG) for further training. We observe some unexpected results that, even if the TG is over-simplified and unrealistic, it drastically accelerates the training. The agent learns a 99\% win-rate on a 64$\times$64 map against the most difficult non-cheating level-7 (L7) built-in AI in 1.08 hours, which can be nearly 100 times faster than previous approaches. We also disclose that the TG is robust and helpful in more situations. It can be used for training agents of different races in SC2, which requires different strategies, and can be adapted. We also show that TG is more effective than previous MB-RL algorithms on SC2. We then present a TG hypothesis that gives the influence of different fidelity levels of TG. 

Real games may have unequal state and action spaces with TG. Transferring policy to it faces many training difficulties. Therefore we propose a novel deep neural networks (DNN) structure for it called XfrNet (``Xfr'' stands for ``transfer'') and validate its usefulness by experiments. Finally, we summarize the useful tricks of training agents on SC2. By applying the above improvements, our agent can achieve good results in the three difficulty levels above 7 (all are cheating difficulty means the built-in AI use cheating to gain a big advantage above our agent). Our winning rates for difficulty 8, 9, 10 reached $95\%$, $94\%$, and $90\%$ respectively \footnote{To facilitate reproducibility, we provide our codes and trained models at \url{https://github.com/liuruoze/Thought-SC2-model-experiment}}.

\begin{figure}[h!]
    \begin{minipage}[t]{\linewidth}
        \centering
        \includegraphics[width=0.95\columnwidth]{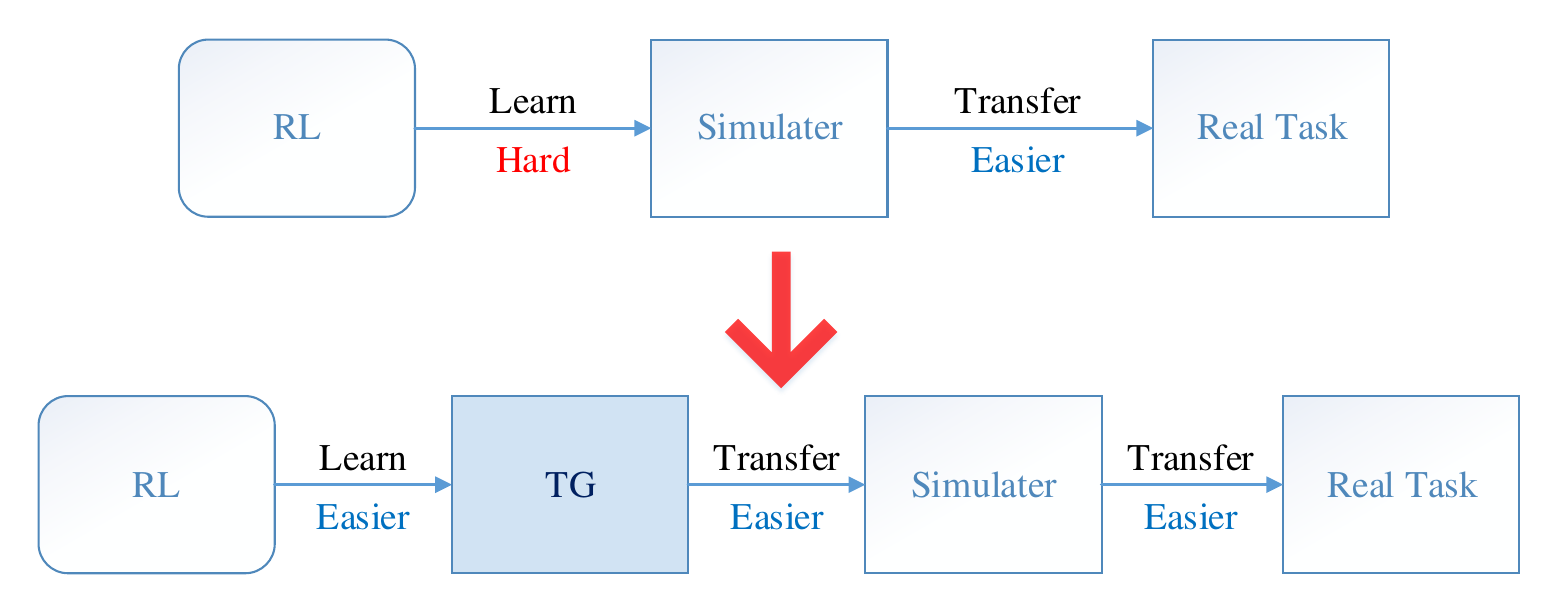}
        \caption{The improved pattern by introducing TG.}
        \label{fig:new pattern}
    \end{minipage}
\end{figure}

We argue that another contribution of TG is to improve the pattern of applying RL in real-world tasks which often follows the below pattern: First, a simulator for the real task is built; Then RL algorithms are applied on the simulator; Finally, the trained policy is transferred to the real task. The motivation for such a process is that simulators are often lightweight and fast. However, the purpose of the simulator is to be as similar to real games as possible, not to facilitate RL training which in contrast, TG was born for RL thus can provide much faster-sampling speed and smoother curriculum difficulty. TG also maintains similarities with the simulator. Therefore, we can first do better RL learning in TG, and then do transfer learning on the simulator by the trained policy in TG which is shown in Fig.~\ref{fig:new pattern}. In short, TG is a lubricant, which bridges the gap between the traditional simulator and RL, which facilitating the RL training on the real task. The main contributions of this paper can be summarized as follows:
\begin{itemize}
\item We propose a TG framework and demonstrate the potential of its effectiveness and fast-speed. The robustness and application of TG on SC2's different races and maps, SC1, and a real-world hydropower task are also tested.
\item A novel XfrNet for TG is proposed to handle the cases of transferring to RG with unequal state and action spaces.
\item We suggest a TG hypothesis that shows the influence of different fidelity levels of TG.
\item TG bridge the gap between the traditional simulator and RL, which facilitate training on the real task.
\end{itemize}

\section{Background}
In this section, we first present the definition of RL and model-based RL. Then we discuss TL and curriculum learning. Finally, previous studies on SC1 and SC2 are given.

\subsection{RL and Model-based RL}
RL handles the continuous decision making problem which can be formulated as a Markov decision process (MDP) which is represented as a 6-tuple $\langle S, A, P(\cdot|s, a), R(\cdot), \gamma, T \rangle$. At each time step $t$ of one episode, the agent observes a state $s_t \in S$, then selects an action $a_t \in A$ accoring to a policy $\pi: s_t \to a_t$. The agent obtains a reward $r_t = R(s_t, a_t)$, and the environment transit to the next state $s_{t+1}\sim P(\cdot|s_t, a_t)$. $T$ is the max time steps for one episode. $G_t=\sum_{k=t}^{T}{\gamma^{k-t} r_{k}}$ is the return while $\gamma$ is a decay discounter. Value function is $V_{\pi}(s_t) = \mathbb{E}_{\pi}(G_t|s_t)$ which is the expected return for $s_t$ while following policy $\pi$. The goal of RL is to get $\text{argmax}_{\pi}\mathbb{E}(V_{\pi}(s_0))$. 

Transition functions $P(\cdot|s, a)$ (also called as dynamic functions) are usually unknown in many MDPs. Therefore, depending on whether the transition function is used, RL can be divided into model-free RL~\cite{mnih2015human,schulman2017proximal,wang2016duelingICML} and MB-RL~\cite{Ha2018worldmodelNIPS,NagabandiKFL2018mbrl}. The advantage of model-free RL is that it is not necessary to learn the transition function, but the disadvantage of it is the low sample efficiency. In contrast, MB-RL methods tend to be more sample efficient, but need a model.

\subsection{TL and curriculum learning}
Transfer learning~\cite{Transfer2010} allows the transferring of knowledge from the source domain to the target domain, assuming similarities between them. TL has been used in RL~\cite{transferRL,transferRLsurvey}. In our study environment, TG is the source domain, and RG is the target domain. For simplicity, we use $A \Rightarrow B$ to represent transferring from $A$ to $B$. For consistency, $\varnothing \Rightarrow A$ represents training from scratch on $A$. Thus, in this work, we have 3 common learning modes which are $\{TG \Rightarrow RG,\varnothing \Rightarrow TG,\varnothing \Rightarrow RG\}$. 

Curriculum learning (CL)~\cite{curriculumLearning, teacherCurriculumLearningRLonNNLS} is a learning process that allows learning to grow incrementally through a series of tasks, e.g., repeatedly transfer learning from easier task to harder task. Pang~\cite{pang2019sc2} used the idea of curriculum learning so that their policy can be optimized smoothly which can be formulated as $RG_{i} \Rightarrow RG_{i+1}$. However, the difficulty levels of the SC2 are not smooth (e.g., L3 is much more difficult than L2), which makes the curriculum not ideal. On the contrary, due to we have control of TG, it is possible to design a smoother curriculum in TG for agent learning.

\subsection{StarCraft I \& II}
Previous research on StarCraft focused on SC1 and can be divided into two settings, full-game, and part-game. Methods on full-game are based on search and planning~\cite{ontanon2013survey} or using RL~\cite{bilibiliSC,FB2SC1}. Some studies~\cite{peng2017multiagent,shaoKun} use RL to solve part-game. Recently, due to the new interface and simulator \textit{pysc2}~\cite{vinyals2017sc2}, SC2 has become a suitable platform for RL research. Training on SC2 face many difficuilties which will be analysed latter. Most previous works tackle the SC2 problem in the way of ``divide and conquer''. E.g., Sun~\cite{Sun2018tsbot} uses Hierarchical RL (HRL)~\cite{dietterich2000hierarchical} and manually designed macro-actions. Pang~\cite{pang2019sc2} uses an HRL architecture combined with learned macro-actions. Lee~\cite{tang2018sc2} handles SC2 with a modular architecture that consists of scripts and machine learning. 

AlphaStar (AS)~\cite{AlphaStarNature} has made a big progress on SC2. It employs a transformer~\cite{attention} and multi-agent self-play RL algorithms. Several differences exits between our work and AS: (1) we use SC2 as an environment for testing TG, while AS use SC2 as a testbed of general RL; (2) AS use human replays to do behavioral cloning and build orders as a guiding reward in the RL training, on the contrary, we use human knowledge to build the forward model; (3) AS is a system consists of many techniques and use huge computing resource while our method focused on 2 techniques and use a single commercial machine.

Some methods are similar to ours. Ontanon~\cite{ontanon2015adversarial} uses abstract models and hierarchical planning in SC1. Unlike them, we use RL and build a model from scratch. Churchill~\cite{SparCraft} analyzed a simulator called \textit{SparCraft} which is similar to the TG. The difference is that: SparCraft is a simulation of the combat module of the game, but the TG abstract the whole game. Ovalle~\cite{BootstrappedMBRL2020} proposed an MB-RL method that learns a multi-headed bootstrapped transition function. They use rules to do error-correcting after prediction, in the way of which share similarities as ours. The difference is that they combine the learning-based and rule-based models. Justesen~\cite{MacroFromReplays2017} introduced a DL method to learn macro-management from replays. It is noted that different from some previous works, we do not use two useful tricks which are: 1. Action-mask, which is to mask the unavailable actions in the action probability layer~\cite{vinyals2017sc2}; 2. Action-dependent, which is when one action needs some precondition, we instead to execute the precondition~\cite{Sun2018tsbot}. Without them, efforts for exploration are increased. In order to verify the improvement of exploration brought by TG, we don't use them. Therefore, the environment we faced is slightly more difficult.

\section{Methodology}
Firstly, we present a discussion and a prototype to illustrate our intuition for our method. Secondly, we give the components of the method. Finally, we introduce the overall method and implementation details.

\subsection{Analysis of RTS games}
We first discuss why RTS games, especially SC2, are difficult for RL. The previous works~\cite{ontanon2013survey,vinyals2017sc2} pointed out that this is caused by four reasons: huge state space, large action space, a long time horizon, and imperfect information. However, these problems may be alleviated like in~\cite{pang2019sc2,Sun2018tsbot,tang2018sc2} by HRL and the design of macro-actions. In this paper, we argue that the reasons may also include three more ones. The first one is that SC2 may not be considered an MDP if historical observations are not used as states, e.g., if we don't store information about the opponents as historical state.

The second reason which is little discussed before is the ``selection problem" causing by pysc2. In order to simulate human's operations, pysc2 needs to first select entities (building, unit, or group) by their positions and then perform actions. The selection includes ``select point" which simulates the click of the mouse and ``select rect" which simulates the mouse's drag box. Without the selection, the actions have no ``subjects'' and thus can not be executed. This brings a challenge for DRL due to the agent must first learn how to accurately select an entity on the game screen or minimap. For this reason, all methods for pysc2 have taken an alternative approach, e.g, some use macro-actions, put the selection in the macro, and the position of the entity is given by raw observations of the game; the other, AlphaStar, makes the arguments of their defined actions include the entity handle which is also given by raw observation. Ideally, we want to use computer vision methods like object detection~\cite{objectDetectionNIPS2013} or segmentation~\cite{ObjectSegDRLinNIPS} to find these entities on images of screens or minimaps, rather than by raw observations which should only be accessed by the game engine. But this way is still difficult for current DRL methods~\cite{ObjectSegDRLinNIPS}.

The third one is the existence of to many rules and dependencies in RTS. Rules mean if a \textit{zealot} needs to be produced, we need a \textit{gateway}, while a gateway requires a \textit{probe} (worker) to build, and probes are usually produced by the main base. Under all preconditions, the zealot still needs 27 seconds to be produced. Dependency refers to the fact that most buildings and units in the game rely on other buildings and technologies. These dependencies form a tree structure called \textit{techtree} in the game (we can consider dependencies as special rules. Below we will use rules to refer to both). These rules constitute a world that requires reasoning to make decisions, which is still difficult for DL~\cite{abductiveLearningNIPS}. Note rules are similar to physical laws in the real world. The neural networks can be trained to get 99.99\% accuracy for some tasks, but its error rate is still 0.01\%, which is essentially different from the zero error rate gained by rules. To compare, there is an Atari game which is called ``Montezuma's Revenge'' (MR). The rules in MR have prompted researchers to spend a lot of effort to explore effective algorithms on it, such as HRL methods. There are only a few rules in MR, while there are hundreds of them in SC2. Because of these, RL algorithms must pay more efforts for exploration to learn them, making training difficult.

\begin{figure*}[h!]
    \centering
    \subfloat[]{
        \centering
        \includegraphics[width=0.475\columnwidth]{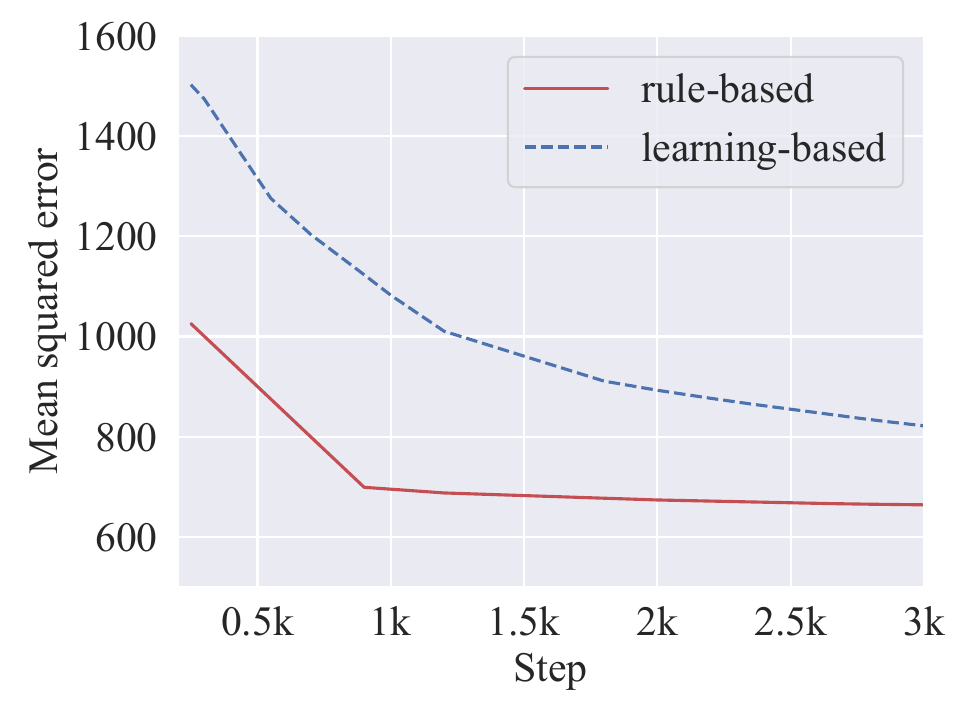}
        \label{fig:s.a.4.2.2}
    }
    % \hspace{0.01cm}
    \subfloat[]{
        \centering
        \includegraphics[width=0.475\columnwidth]{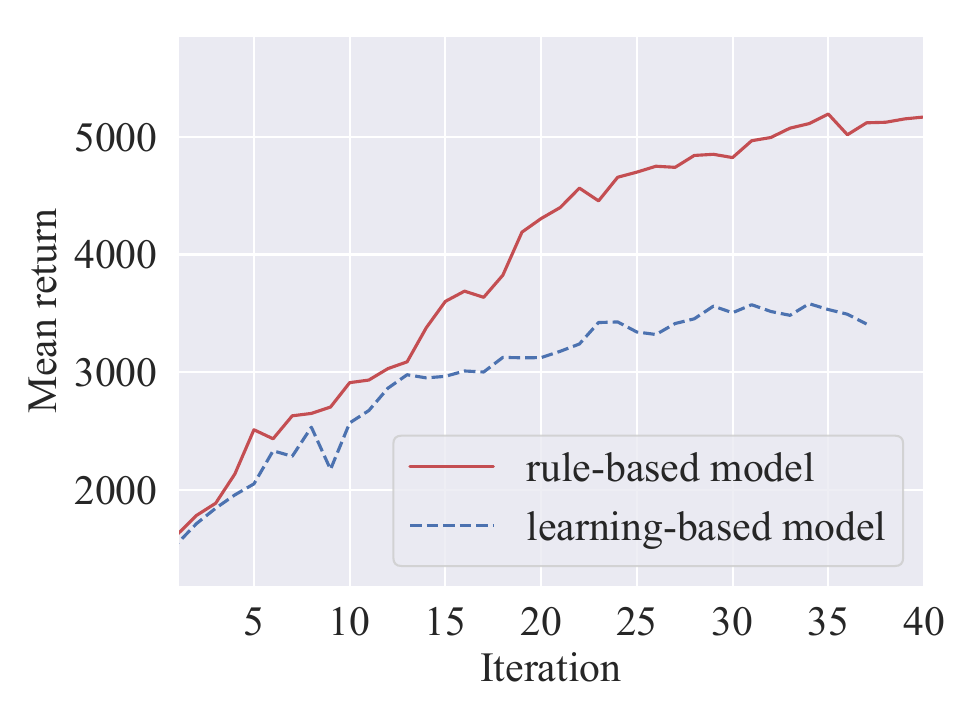}
        \label{fig:s.a.4.2.1}
    }
    % \hspace{0.01cm}
    \subfloat[]{
        \centering
        \includegraphics[width=0.475\columnwidth]{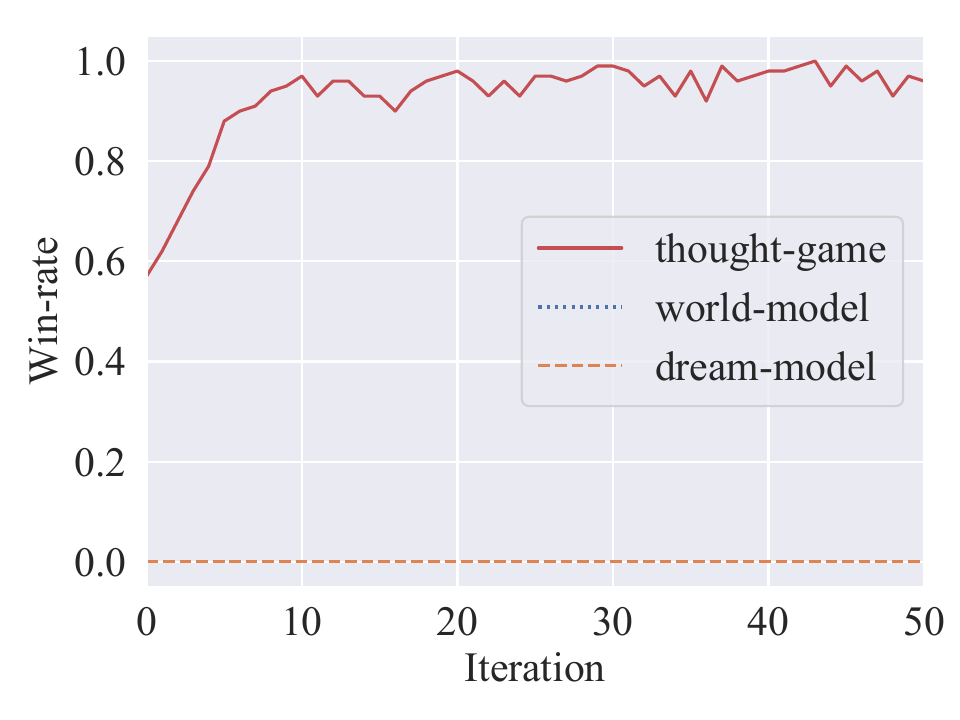}
        \label{fig:s.a.4.2.3}
    }
    % \hspace{0.01cm}
    \subfloat[]{
        \centering
        \includegraphics[width=0.475\columnwidth]{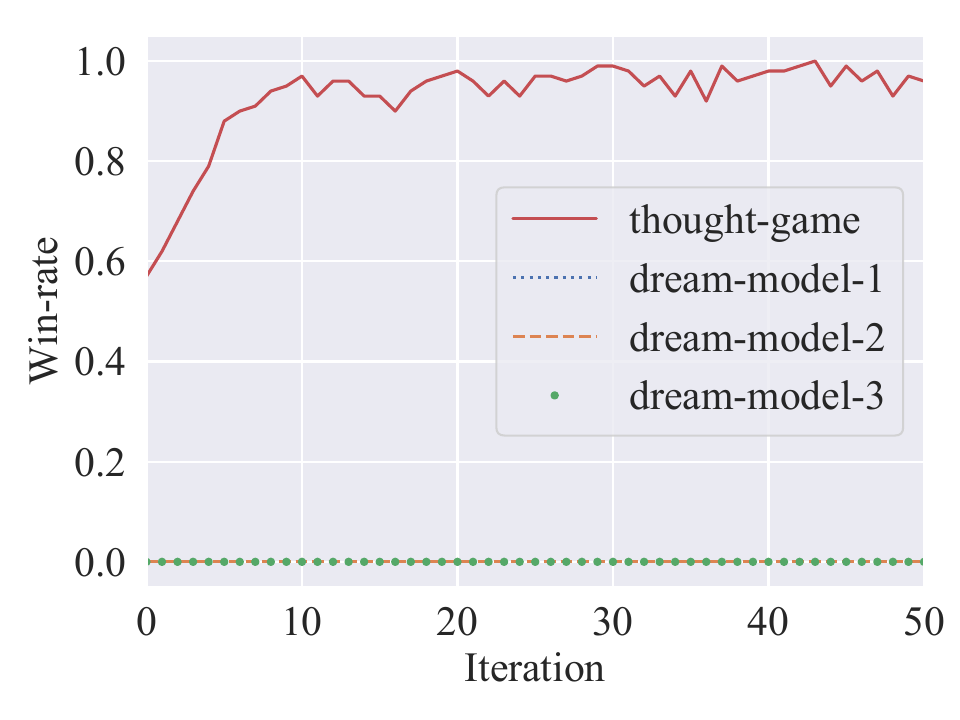}
        \label{fig:s.a.4.2.4}
    }
    \caption{(a)~Training process of the learning-based and rule-based models. (b)~Effects of using the learning-based and rule-based model to learn policy in PT. (c)~Effects of training in RG using TG, world-model, and dream-model. (d)~Effects of training in RG using TG and different iteration of dream-models. PT=Prototype. TG=thought-game. RG=real game.}
    \label{fig:s.a.4.2.a}
\end{figure*}

\subsection{Prototype}
If we can obtain a dynamic model of the environment, the efficiency of RL may be improved. Like many MB-RL works, we first try to learn a model using learning methods such as in~\cite{NagabandiKFL2018mbrl}. We build a prototype and then compare the learning-based model with a rule-based model on it. The learning-based method~\cite{NagabandiKFL2018mbrl} is worked as follows. First, it collects numerous trajectories that consisting of $\{s_t, a_t, s_{t+1}\}$ using random policy. Second, it uses the trajectories to learn a dynamic function as $s'_{t+1}= f_\rho(s_t,a_t)$. Third, it uses the dynamic to generate fake trajectories, combined them with real trajectories, and then learn a policy on them. The prototype is a task in a full game of SC2 - development of economy which is to maximize the number of minerals owned by the agent in a fixed period of time. The agent has three actions: training workers, building \textit{pylons}, and doing nothing. Specifically, adding workers will increase the collection rate of minerals but consume $50$ minerals and $1$ food supplies. Building pylon will increase food supplies by $8$ and consume $100$ minerals. The ideal number of workers for collecting is limited. When this limit is exceeded, adding workers will not increase the collection rate of minerals. Therefore, the agent needs to properly control the three actions.

We learn a model by minimizing the MSE (mean squared error): $\sum(s'_{t+1} - s_{t+1})^2$. After that, we calculated the H-step-error~\cite{NagabandiKFL2018mbrl} of the model which is $\frac{1}{H}\sum_{z=1}^{H}\|s'_{t+z}-s_{t+z}\|^{2} $ where $s'_{t+z}= f_{\rho}(s'_{t+z-1},a_{t+z-1})$ and found it is very large when $H$ is $5$. We deduce the learned model is not so accurate. The reason may be as follow: 1. the dynamic is various and the random policy doesn't capture all circumstances; 2. the model of the game is fragile and sensitive, thus any errors in the model make a big compound error. On the contrary, we can use the rules of SC2 to predict some features of the next state accurately (this has a precondition that SC2 is a deterministic game, so we can use rules to build that). We find that the new rule-based model learns much faster, as shown in Fig.~\ref{fig:s.a.4.2.a}(a). We also found the H-step-error of the new model is smaller. Fig.~\ref{fig:s.a.4.2.a}(b) shows the policy trained on the new model is faster than the learning-based one. This experiment gives us an intuition that if we can use known rules to build a dynamic model, then it may help RL better than a learned one.

\subsection{Thought-game}
Suppose we extend the environment to a full game of SC2, the operation of implementing all number of rules may be very costly. In this work, we introduce a simplified and abstract model, through which we can learn an initial policy by RL and then transfer it to the real game by TL. A rule-based model TG is built to abstract the RG. Concretely, a definition of TG based on MDP is shown below. Suppose RG is $M_s=\langle S_s, A_s, P_s(\cdot), R_s(\cdot), \gamma, T_s \rangle$. We use subscript $s$ (means source) for RG and subscript $m$ (means mini) for TG throughout the paper. We want the optimal policy $\pi_{\theta}$ on $M_s$. TG is $M_m=\langle S_m, A_m, P_m(\cdot), R_m(\cdot), \gamma, T_m \rangle$ and a policy $\pi_{\phi}$ exists in TG while $a_m=\pi_{\phi}(s_m)$. If a state mapping function $f_s: S_s \to S_m$ and an action mapping function $f_a: A_m \to A_s$ exist, then an action in RG can be obtained:
\begin{equation}
    a_s = f_a(a_m)=f_a(\pi_\phi(s_m))=f_a(\pi_\phi(f_s(s_s)))
    \label{eq02}
\end{equation}
Due to $a_s=\pi_\theta(s_s)$, the equation~\ref{eq02} can also be written as $\pi_\theta(s_s)=f_a(\pi_\phi(f_s(s_s)))$. The architecture is shown in the left of Fig.~\ref{fig:Thought-Game}. If we have $f_s(\cdot)$, $f_a(\cdot)$ and the policy $\pi_{\phi}$, we get $\pi_{\theta}$ easily. However, trying to calculate $f_s(\cdot)$ and $f_a(\cdot)$ is intractable. To make it practical, we do a simplification. We let $\varphi(S_m) \subset \varphi(S_s)$, in which $\varphi(\cdot)$ denotes the features of the state, and $A_m=A_s$, e.g., use macro-actions in the RG and ensure actions in the TG have the same semantics. In the~\ref{section:xfrnet} XfrNet section, we will show how to handle the case when $S_m \neq S_s, A_m \neq A_s$. Hence, we redefine TG as:
\begin{align}
  M_m=\langle& \varphi(S_m) \subset \varphi(S_s), A_m=A_s, \nonumber \\
  &P_m(\cdot) \approx P_s(\cdot), R_m(\cdot), \gamma, T_m\rangle
\end{align}
We design $P_m(\cdot) \approx P_s(\cdot)$ and will show how to do it later. $R_m(\cdot)$ use the outcome reward. $T_m$ is the max time step for TG. After designing TG, we can use model-free RL or MB-RL to learn an optimal policy on it. Once the policy $\pi_{\phi}$ is obtained, we can operate the agent in the RG through the mapping function or has a good initial policy to train the policy $\pi_{\theta}$ in the RG. 

\begin{figure}[h!]
    \begin{minipage}[t]{\linewidth}
        \centering
        \includegraphics[width=0.75\columnwidth]{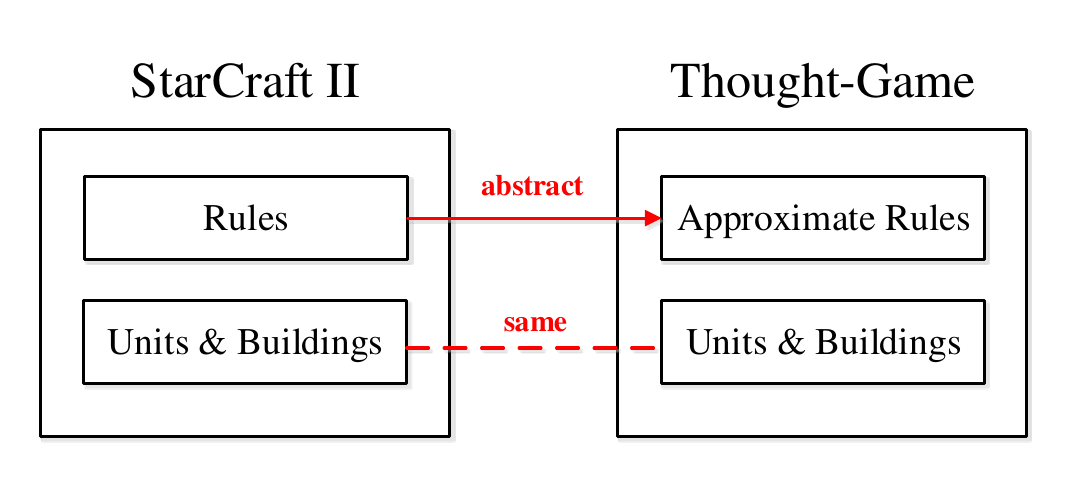}
        \caption{A module description of TG-SC2.}
        \label{fig:thought sc}
    \end{minipage}
\end{figure}

\begin{figure*}[h!]
    \begin{minipage}[t]{\linewidth}
        \centering
        \includegraphics[width=0.95\columnwidth]{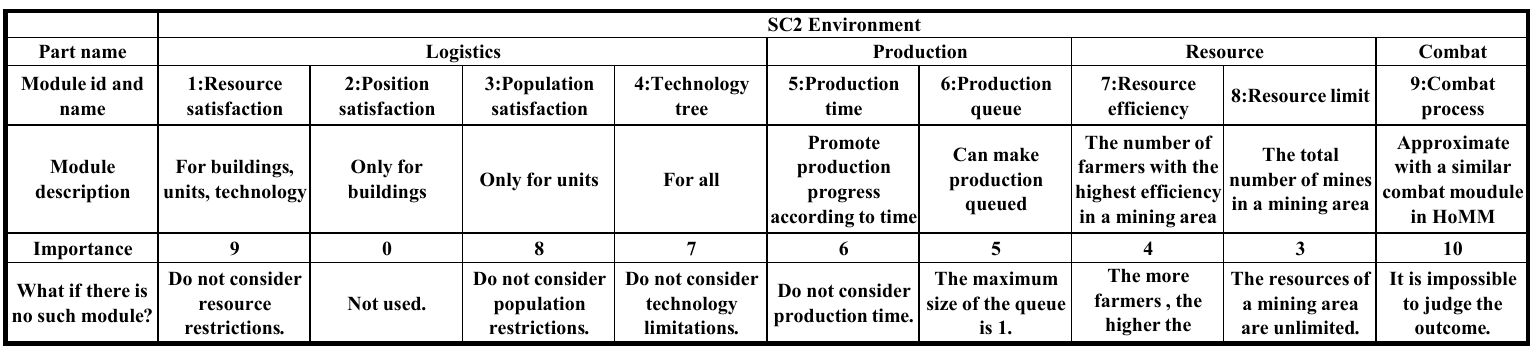}
        \caption{Modules of StarCraft II.}
       \label{fig:modules}
    \end{minipage}
\end{figure*}

\begin{figure*}[h!]
    \begin{minipage}[t]{\linewidth}
        \centering
        \includegraphics[width=0.95\columnwidth]{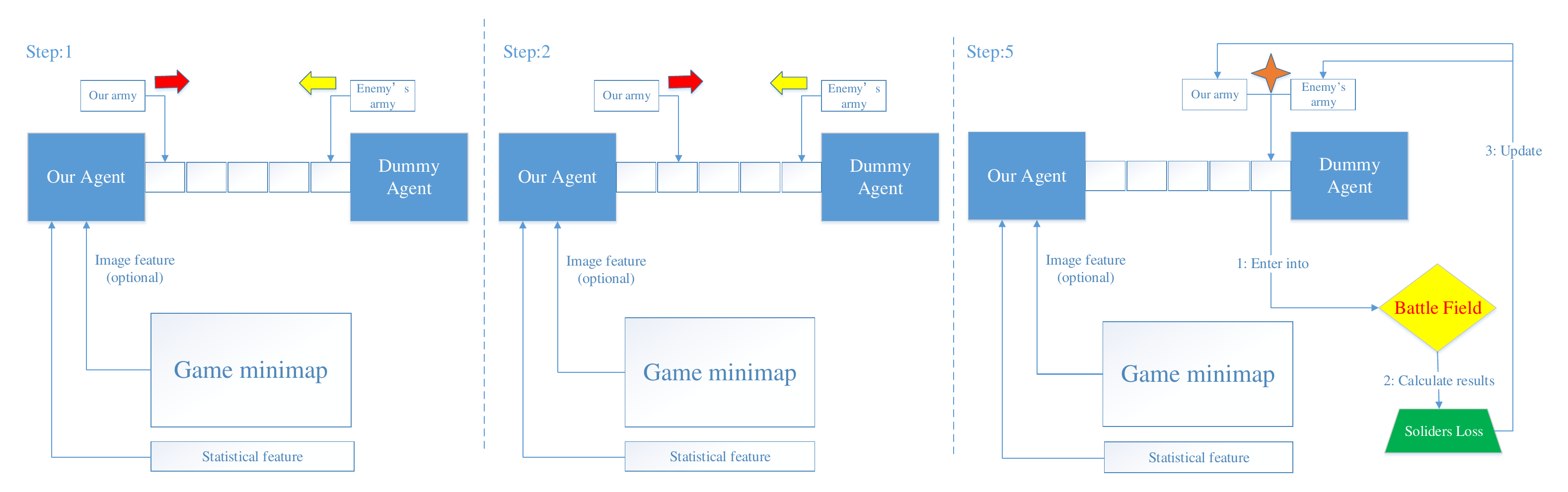}
        \caption{A spatial-temporal description of the battle process in TG-SC2. In terms of spatial, as we are playing 1v1 mode, we assume the main-bases of both players are in a straight line. The length of the line represents the distance between the two players in the RG map. This value can be set (e.g., 5 in this experiment). Each player has an army (simulating the control group in the RG, that is, different types of soldiers) in his base. When the player commands ``attack", its army begins to move to the enemy's base, one grid per one step. When two armies meet in the same grid, they will enter a battlefield on which each army calculates total damages (enemy's loss) based on its soldiers' damage, number, and ranged type. At the end of the step, the army makes a reduction based on the loss (melee first). Then the next step. If one's army is at the opponent's base and the opponent's soldiers reduce to empty, the excess damage is allocated to buildings. If the hit points of all buildings of one side drop to 0, it loses the game. In our experiment, 1 step in TG simulates $9$ seconds of RG.}
        \label{fig:TG step}
    \end{minipage}
\end{figure*}

\subsection{Model design} \label{section:TG design}
We now explain how to design $P_m(\cdot) \approx P_s(\cdot)$. Take SC2 as an example, the design of TG can be divided into two parts, one part is the units and buildings, and the other is the rules of SC2 (which is shown in Fig.~\ref{fig:thought sc}). The former can be got from a public wiki website of SC2, we then turn them as classes in object-oriented languages (e.g., Python). The latter should be abstracted from SC2. The abstract process is as follows. Firstly, we divide the rules of SC2 into 9 modules (see Fig.~\ref{fig:modules}). Secondly, we abstract these modules one by one according to their importance. Some rules have parameters, e.g., if workers increase the minerals by collecting resources, how many minerals does a worker collect in each time step of TG? Note that a time step in TG is not identical to that in SC2. These values are given in a heuristic manner, like $5$ in this setting. For complex module like the combat module, we abstract it as a tabular turn-based strategy module in the well-known game \textit{Heroes of Might and Magic}. The combat module is detailed described in Fig.~\ref{fig:TG step}. The motivation of this design is that the battle design of most games comes from the real-world so that some rules can be shared. The principle of abstraction is that: abstract the modules in the RG with similar rules and heuristically chose rule-parameters while ensuring that the actions in both games have similar semantics. The ``similar semantic'' means that the distance (which can be formulated as KL divergence) between the trained policy on TG and the one on RG is shorter than that between the random policy and the one on RG. We will validate it through experiments later. Based on the principle, the GTG (Generate a TG) process can be described as:
\begin{enumerate}
    \item Divide the rules of RG into several modules and sort them by the importance;
    \item Abstract these modules into TG according to their importance. The most important one being the first;
    \item Use heuristics to decide the value of the rule-parameters of these modules;
    \item If some modules in RG are complex, use simpler modules of another RG to imitate them.
\end{enumerate}
In fact, for RTS games that involve built-in AI, the training difficulties come from two parts. The first is how the agent learns to interact with the world. The second is the strength of the opponent's built-in AI. If we consider the level of built-in AI as part of the environment. Then the RG of SC2 can be seen as a set $\{\text{RG}_i\}$, ($i$ is the level and $1 \le i \le 10)$. We can also design a curriculum for TG by building different opponents in the TG as $\{\text{TG}_j\}, (1 \le j \le 7)$ and the details are in section~\ref{section:details}.

\subsection{Training algorithm}
We now give a training algorithm based on TG. First, we use the ACRL algorithm to train an agent in the TG from scratch. The idea of the ACRL algorithm is simple, i.e., automatically increase the difficulty level of the curriculum when monitoring that the win-rate exceeds a threshold (see Algorithm~\ref{alg:acrl}). We will show the effectiveness of ACRL in the experiment section. In this work, we focus on using TL methods. After we train a policy in TG, we map this policy to RG for continued training using TL algorithms like finetune~\cite{Finetune2014}, a widely-used TL technology in DL. 

We use a DNN as our policy model. It contains an input (observation) layer and two hidden layers which are called layer\_1 and layer\_2. The action logit layer and value layer are after the layer\_2. The value layer outputs the state's value function. The action logit layer is then going to a softmax layer to output probabilities for each action. This network is called OrigNet, which is distinguished from the next proposed architecture.

\subsection{TG hypothesis}
Since the main purpose of the agent learning in TG is to be familiar with the rules that have migrated from RG. This can be summarized as a TG hypothesis. \textbf{Thought-game hypothesis:} \textit{The fewer rules that are migrated from RG to TG, the easier it is for an agent to learn in TG, and it is harder to transfer to learn in RG; The more rules that are migrated from RG to TG, the harder it is for an agent to learn in TG, and it is easier to transfer to learn in RG}. TG hypothesis argues a trade-off between learning speed in TG and transferred performance in RG. We will use experiments to validate it. Suppose RG has $n$ parts and TG has a property \textit{level}. When the level is $n$, it means TG has the maximum level of abstraction (we refer it to TG$^n$), containing the least but most important rules in RG. TG$^0$ contains all the rules in RG, which has the least abstraction level. Based on them, the pseudocodes of TG is given below:
\begin{lstlisting}[style = python]
class TG:
    def __init__(self, level):
        self.level = level
    def step(self, action):
        if self.level < n:
            f = 1stRules()
            self.state = f(self.state, action)
        if self.level < n-1:
            f = 2ndRules()
            self.state = f(self.state, action)
        # More rules
        return self.state
\end{lstlisting}

\begin{figure*}[h!]
    \begin{minipage}[t]{\linewidth}
        \centering
        \includegraphics[width=0.90\columnwidth]{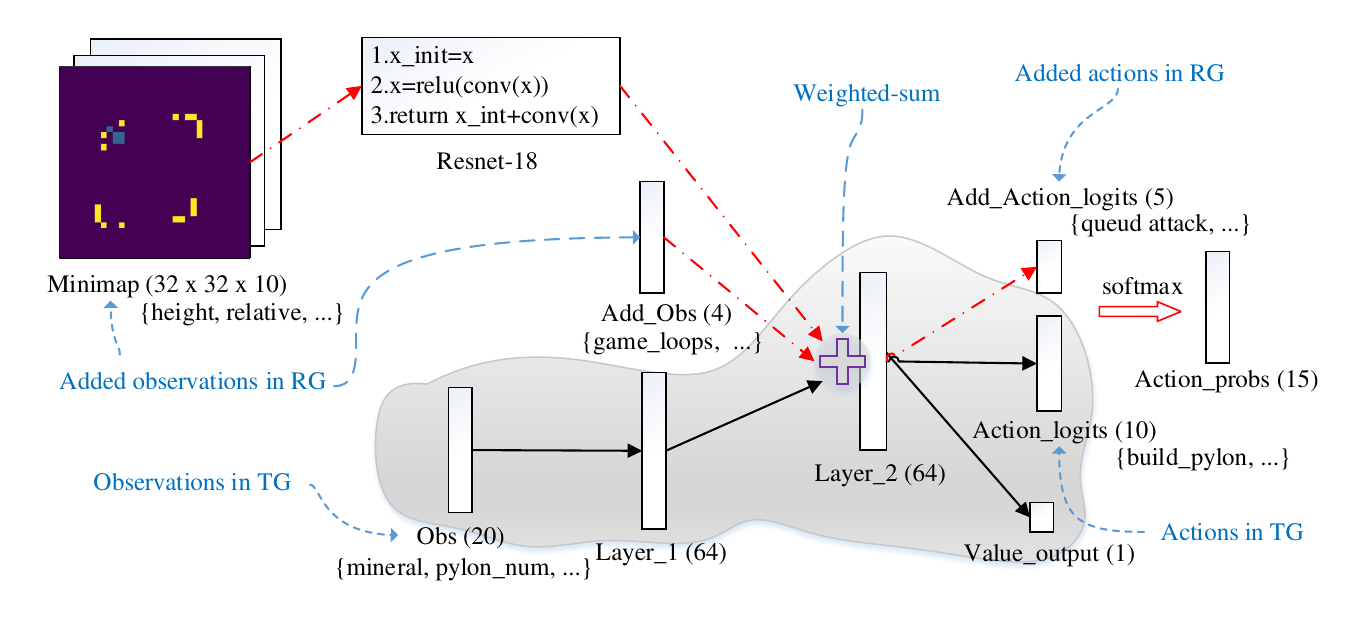}
        \caption{The proposed XfrNet. The layers in the grey region are OrigNet inside XfrNet. The value in the parenthesis after each layer name is the number of units for that layer. The text in the braces besides each layer name is the sample input features for that layer. Solid black lines are weighs trained in $\varnothing \Rightarrow TG$ and then be restored in $TG \Rightarrow RG$. Red dash lines are weights newly trained in $TG \Rightarrow RG$. We use the minimap as the image input and pass it through a Resnet-18 (meaning it has 18 Resblocks). We use a weighted sum of the added image input, added observation, and the output of the Layer\_1. Moreover, we only partially restore the weights in the action logit layer.}
        \label{fig:add network}
    \end{minipage}
\end{figure*}

\subsection{XfrNet} \label{section:xfrnet}
We often want to get more input information and the action space may be different from TG to RG. Hence, we propose a novel DNN architecture for TL in the RL domain called XfrNet. XfrNet is an extension of OrigNet. An add\_obs layer plus minimap layer are added to handle the added observations in RG. Minimap information is processed by a ResNet-18. An add\_action\_logits layer is also added after layer\_2 to handle the added actions in RG. The description of XfrNet and OrigNet can be seen in Fig.~\ref{fig:add network}. The XfrNet can be trained in TG and then be transferred to RG under the conditions that $S_m \leqslant S_s$ and $A_m \leqslant A_s$ (this has the same meaning as $S_m \subset S_s$ and $A_m \subset A_s$. We use $ \leqslant $ for a more intuitive representation). Note that XfrNet can also handle the case of $S_m > S_s$ or $A_m > A_s$ of which are trivial by using state padding and action masking.

The advantage of XfrNet is as follows: 1. It can use additional input information like time or minimap; 2. It can utilize additional actions in the RG; 3. Based on the previous two, the training is still fast. We also use finetune to train XfrNet. The normal mode of finetune is to freeze the trained previous layers, and train newly added layers by a small learning rate. We found that, due to the difference between TG and RG, if we use freezing, the result is not ideal. Several other important mechanisms we proposed for training XfrNet are shown below and the validatiton results are presented in the experiment section.

\subsubsection{Partially restore} 
As a result of the added actions in the output layer, the shape of the weight in the action logit layer is different. Therefore, we can not directly restore that layer's weight. To handle this problem, our solution is \textit{partially restore}, which is only restoring parts of the weights of the action logit layer. Then the data in that layer will pass to the softmax layer to generate probabilities of all actions.

\subsubsection{Weighted sum} 
An intractable point of TL in XfrNet is how to use the newly added observations while retaining previous trained experience. We make the newly added observation has additional routes and then be merged into one exchange layer. To handle the problem of the impact of newly initialed weights, we choose a weighted sum addition for the merging operation which is as follow:
$o = (1- w) o_{1} + \frac{1}{2} w \cdot o_{2} + \frac{1}{2} w \cdot o_{3}$ where $o_{1}$ is the output of Layer\_1 and $o_{2}$ is the output of Add\_obs and $o_{2}$ is the output of Resnet-18 while $w$ is the weight paramter.

\subsubsection{Customize initialization} 
For the weights which are not restored, the initialization strategy will affect the final learning effect. We tried widely-used Xavier~\cite{XavierInitial} and He~\cite{HeInitial} initialization and found their results are not satisfying. We present a simple initialization strategy which is $w_{init} = U(-d, +d)$ and $ b_{init} = U(-d, +d)$ where $U$ means random uniform distribution where $d$ is a hyper-parameter. We found this strategy has a better learning effect in our experiment.

\subsection{Complete method}
\begin{algorithm}[th]
    \caption{Training with Thought-game (TTG)}
    \label{alg:train}
    \textbf{Input}: RG and its difficult levels: $\{\text{RG}_i\}$, state and action space $S_s$ and $A_s$.
    \begin{algorithmic}[1] %[1] enables line numbers
        \STATE Design the forward model TG (state space $S_m$ and action space $A_m$) acroding to section~\ref{section:TG design} and each difficult levels of it, denoted as $\{\text{TG}_j\}$
        \STATE Run ALG.~\ref{alg:acrl} ($\tau=\text{TG}, \alpha=1, \beta=7, \pi=\pi_{\phi}, \eta=0.95$)
        \STATE Transfer learning from $\pi_{\phi}$ on $\text{TG}_7$ to $\pi_{\theta}$ on $\text{RG}_7$, using XfrNet if $S_m \not= S_s$ or $A_m \not= A_s$
        \STATE Run ALG.~\ref{alg:acrl} ($\tau=\text{RG}, \alpha=7, \beta=10, \pi=\pi_{\theta}, \eta=0.93$)
    \end{algorithmic}
\end{algorithm}

\begin{algorithm}[th]
    \caption{Automated Curriculum RL (ACRL)}
    \label{alg:acrl}
    \textbf{Parameter}: environment $\tau$, threshold $\eta$, begin level $\alpha$, end level $\beta$, policy $\pi$.
    \begin{algorithmic}[1]
    \FOR {level $=\alpha$ to $\beta$}
        \WHILE {win-rate $\omega < \eta$}
            \STATE Run ALG.~\ref{alg:ppo} in $\tau_i$ to improve policy $\pi$
            \STATE Compute $\omega$ for $\pi$ in $\tau_i$
        \ENDWHILE
    \ENDFOR
    \end{algorithmic}
\end{algorithm}

\begin{algorithm}[th]
    \caption{PPO algorithm we used}
    \label{alg:ppo}
    \begin{algorithmic}[1]
    \FOR {$\text{i}=1$ to $500$}
        \FOR {$\text{actor}=1$ to $100$} 
            \STATE Run for $T$ timesteps, compute $\hat{A}_1, \dots, \hat{A}_T$
            \STATE Minimize $L_{p}$, with $10$ epoches, $256$ batch-size 
            \STATE $\pi_{old} \leftarrow \pi$
        \ENDFOR
    \ENDFOR
    \end{algorithmic}
\end{algorithm}

The overall method for SC2 can be summarized as simple as follows: 1.Design TG and its different levels. 2.Use CL to train a policy on TG (from TG$_1$ to TG$_7$). 3.Use TL to transfer the policy to RG$_7$ with XfrNet. 4.Use CL to train the policy on RG (from RG$_7$ to RG$_10$). The detailed algorithm can be seen in Algorithm~\ref{alg:train}. The RL algorithm in ACRL we used is $\epsilon$-cliped version of PPO~\cite{schulman2017proximal} (see Algorithm~\ref{alg:ppo}). The PPO loss is defined as : $L_{p} = L_{c} + c_1 L_{v} - c_2 L_{e}$ in which are:
\begin{align}
    & L_{c} = -\mathbb{\hat{E}}_t [\text{min}(\text{clip}(\rho_t, 1-\epsilon, 1+\epsilon) \hat{A}_t, \rho_t \hat{A}_t)] \\
    & L_{v} = \mathbb{\hat{E}}_t [(r_t + \gamma V_{\pi}(s_{t+1})- V_{\pi}(s_t))^2] \\
    & L_{e} = -\mathbb{\hat{E}}_t [\pi(\cdot| s_t) \cdot \text{log}(\pi(\cdot| s_t))]
\end{align}
where $\rho_t = \frac{\pi(a_t | s_t)}{\pi_{old}(a_t | s_t)}$. $\hat{A}_t$ is GAE (general advantage estimater): $\delta_t + (\gamma\lambda)\delta_{t+1} + \cdots + (\gamma\lambda)^{T-t-1}\delta_{T-1}$ where $\delta_t = r_t + \gamma V_{\pi}(s_{t+1})- V_{\pi}(s_t)$. The reason why we do not directly transfer the policy to RG$_{10}$ is due to follow. RG$_8$ to RG$_{10}$ are cheating level AI, making their strategies more random. Through experiments, we find that direct transferring has a worse training effect than the method presented here.

\subsection{Implementation details} \label{section:details}
The opponent in TG is named Dummy. Dummy has 5 buildings at the start, and its soldiers are automatically increased over time (note its soldiers are not produced from buildings). When its number of soldiers exceeds a value (50 in our setting), the dummy will attack our agent. We can control the difficulty of TG by adjusting the amount or frequency of its added soldiers to achieve a smoother curriculum. The details of the design of different levels of dummy show in the pseudocodes below. Though we have 7 levels for dummy, we actually use 4 different levels (L1 is equal to L2 and so on), and we found it performs adequate well in practice.
\begin{lstlisting}[style = python]
class Dummy:
    def __init__(self, diff):
        self.diff = diff
    def step(self, action):
        steps = self.steps
        add = self.add_unit
        if steps % 5 == 1:
            add(Terran.Marine(), 1)
            if steps > 15 and self.diff >= 3:
                add(Terran.Marine(), 1)
            if steps > 25 and self.diff >= 5:
                add(Terran.Marine(), 1)
            if steps > 35 and self.diff >= 7:
                add(Terran.Marine(), 2)
    # Other functions
\end{lstlisting}

%if self.military_num() > 50:
%self.army.order = Order.ATTACK

For XfrNet, the number for units in each layer is described in Fig.~\ref{fig:add network}. The weight parameter $w$ is set to $0.2$. The hyper-parameter $d$ is set to $1.0 / 500$. The desgin of state and action space are provided in section~\ref{section: state and action}. We do not use any hand-designed rewards, that is, in the final step, the reward is 1 for the victory, -1 for the loss, and 0 for the draw, and reward is 0 in all other steps. The clip-value $\epsilon$ in PPO is $0.2$. $c_1 = 0.5, c_2 = 10^{-3}, \gamma = 0.9995$. The $\lambda$ is set to 0.9995. Learning rate is set to $10^{-4}$.

We use a multi-process and multi-thread setting to accelerate training. Specifically, we use 10 processes and 5 threads to sample data. Some other important training details and tricks are analysed in section~\ref{section: tricks}. The TG-SC2 has 7 python files, 4 of which are ``agent'', ``mini\_agent'', ``units'', ``strategy\_env'', and the other 3 are ``protoss\_unit'', ``terran\_unit'', ``zerg\_unit'' which are python object files for each race's buildings and units. There are a total of 923 lines of code (without the info for buildings and units). Our 3 programmers and 1 expert use 3 days to build the TG-SC2.

\section{Experiments}
We first provide the setups for our experiments. Then we give the experiments on SC2. After that, the experiments on SC1 and a hydropower task are provided. Finally, we give the useful tricks we found in the training.

\begin{figure}[t]
    \begin{minipage}[t]{\linewidth}
        \centering
        \includegraphics[width=0.95\columnwidth]{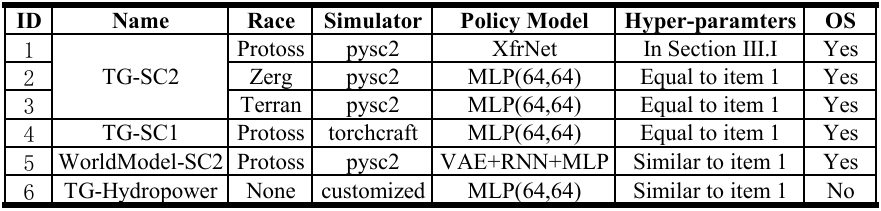}
        \caption{Setups for all experiments. OS=Open-Sourced.}
        \label{fig:experiment suite}
    \end{minipage}
\end{figure}

\subsection{Setups}
As described in Fig.~\ref{fig:experiment suite}, we have done experiments on TG-SC2, Tg-Hypothesis, TG-SC1, WorldModel-SC2, and TG-Hydropower. The agent race, simulator, policy model, and hyper-parameters are outlined in the table. All experimental codes are open-sourced in our codebase for reproducibility except TG-Hydropower due to it uses some real parameter that must be confidential. All our experiments are done at one normal machine. The CPU is Intel(R) Xeon(R) Gold 6248 CPU 2.50GHz with 48 cores. The memory is 400G. The GPUs are 8 NVIDIA Tesla V100 32G, but we only use 4 of them.

In order to be consistent with the previous work~\cite{pang2019sc2} on SC2 and keep the settings simple, we use a full-length game setting, that is, we let the agent use only the first two combat units and do not build any expansion base. We use the same macro-action learned in~\cite{pang2019sc2}. The agent choose a macro-action every two seconds. The actions for combating are: select the army then attack point (enemy base); select the army then move point (our base); do nothing. We make the random seed of the pysc2 in SC2 fixed, thus the locations of our base and enemy base are fixed. The location in RG is represented as the position on the map. In TG, it is represented as the position of the grid. Because the position in the macro-action is conceptual, e.g., we don't need to specify the actual position. So actions in TG can be set as equal to RG. Using XfrNet, our agent can use more actions and attack more locations. 

\subsection{Effect of training}
First, we train a \textit{Protoss} agent using the ACRL algorithm from the easiest level to the hardest level in TG. To test the effectiveness of ACRL, we run experiments with ACRL 5 times and ones without it 5 times and calculate the improvement by the mean and the \textit{p-value}. We find an average improvement of $7.5$ win-rate percent point by using ACRL and the p-value is $0.0225$. After the agent was trained in the TG, we observe its performance. We found that the agent will follow an action flow of ``collecting resources, then building construction, then producing soldiers, then attack”. This shows that the agent has basically learned how to win a victory in the game. 

\begin{figure*}[h!]
    \centering
    \subfloat[]{
        \centering
        \includegraphics[width=0.475\columnwidth]{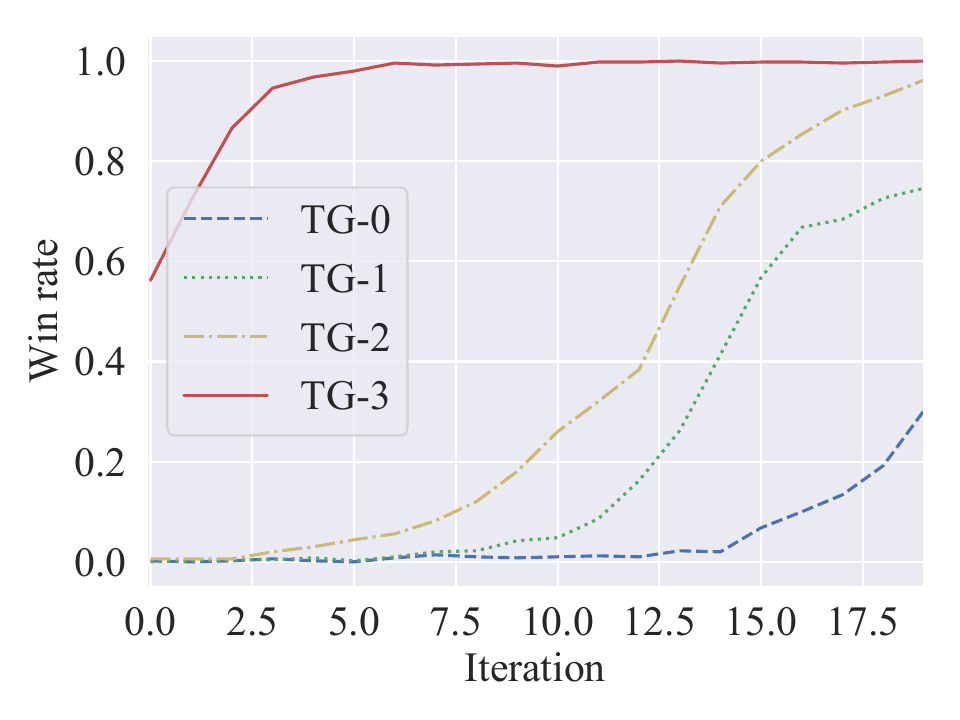}
        \label{fig:s.2.3.1}
    }
    % \hspace{0.01cm}
    \subfloat[]{
        \centering
        \includegraphics[width=0.475\columnwidth]{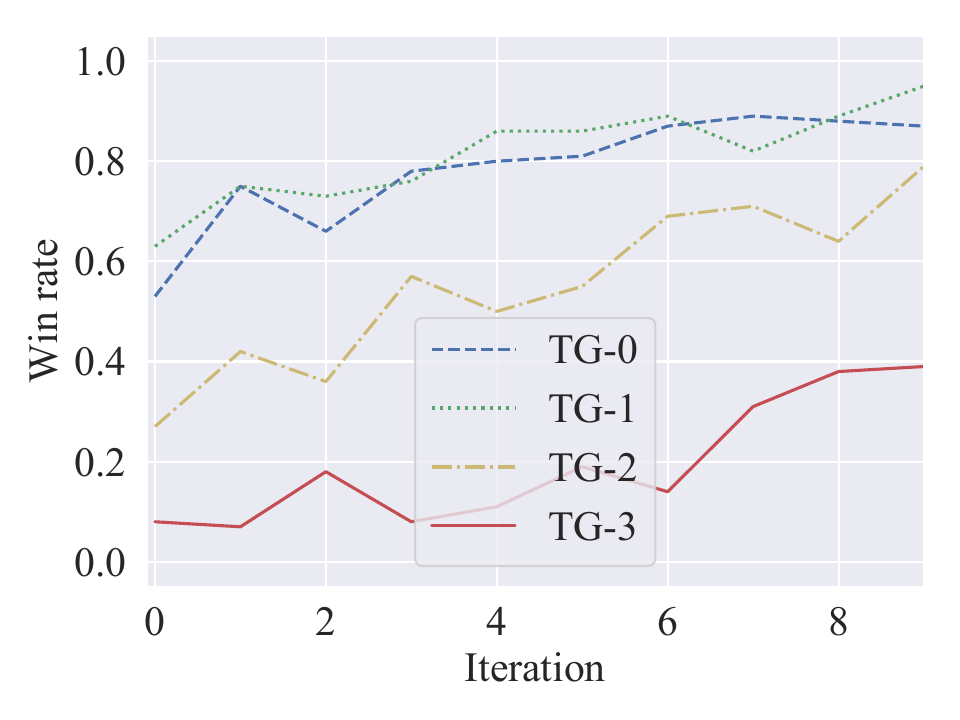}
        \label{fig:s.2.4.2}
    }
    % \hspace{0.01cm}
    \subfloat[]{
        \centering
        \includegraphics[width=0.475\columnwidth]{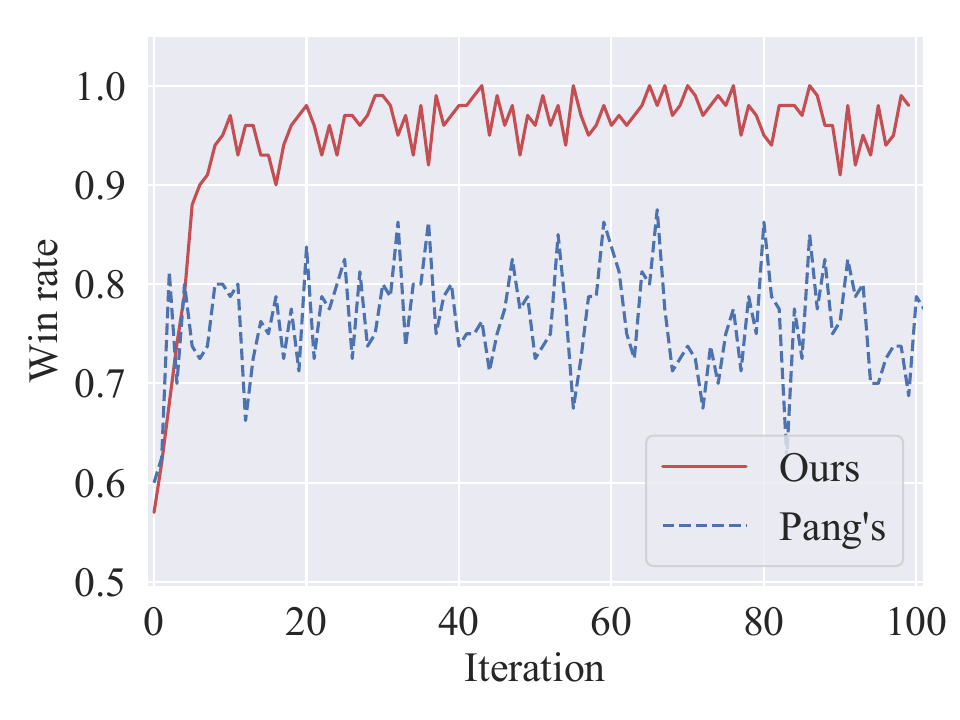}
        \label{fig:s.2.5.1}
    }
    % \hspace{0.01cm}
    \subfloat[]{
        \centering
        \includegraphics[width=0.475\columnwidth]{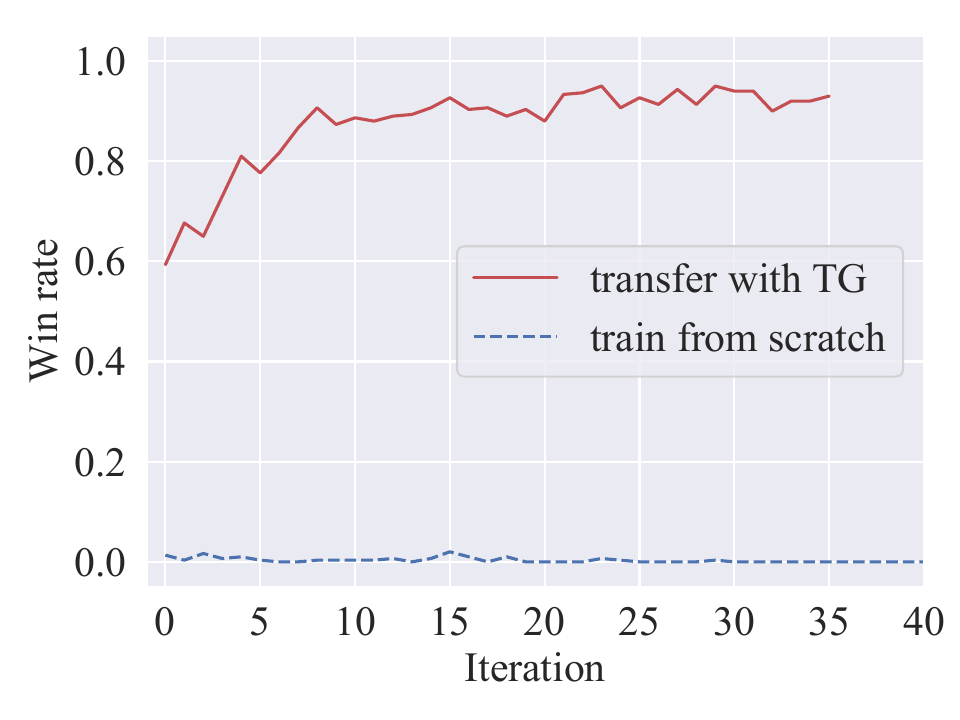}
        \label{fig:s.2.6.2}
    }
    \caption{(a)~Learning in different levels of TGs. (b)~Win-rate of learning in RG from training 30 iterations firstly in different TGs. (c)~Ours vs. Pang's while $TG \Rightarrow RG$ on map \emph{Simple64}. (d)~$TG \Rightarrow RG$ vs. $\varnothing \Rightarrow RG$ on SC2.}
    \label{fig:s.a.1}
\end{figure*}

Then we transfer the trained policy to RG to train our agent against an L-7 \textit{Terran} bot. As shown in Fig.~\ref{fig:s.a.1} (c), our win-rate starts at around $0.5$, and then goes to nearly $1$. The change is when the agent produces the first soldier, it is more intended to attack the enemy's base instead of hoarding it at home. This brings a tactical style that suppresses the enemy which is called \textit{Rush} of which effectiveness is confirmed on many bots~\cite{ontanon2013survey}.
\begin{figure*}[h!]
    \centering
    \subfloat[]{
        \centering
        \includegraphics[width=0.475\columnwidth]{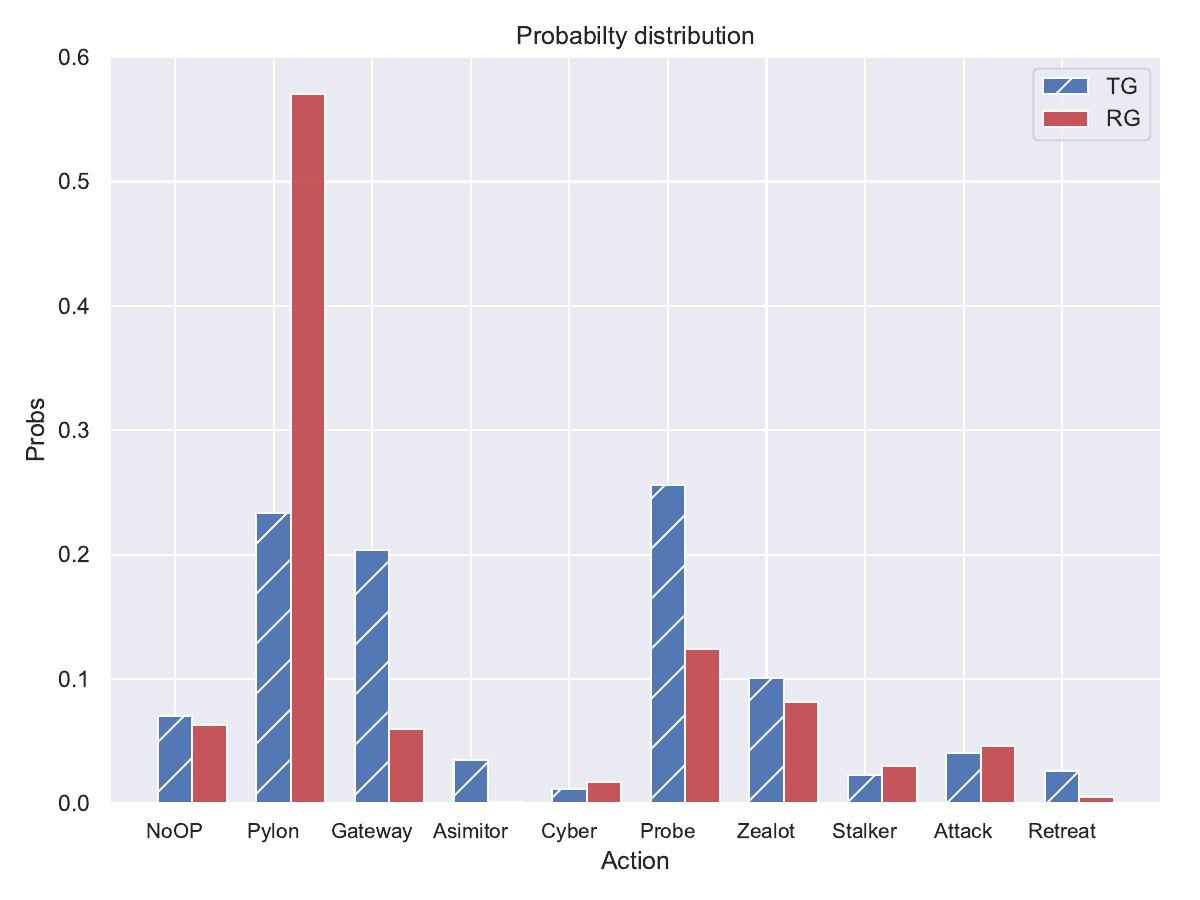}
        \label{fig:s.b.4.4.1}
   }
    \subfloat[]{
        \centering
        \includegraphics[width=0.475\columnwidth]{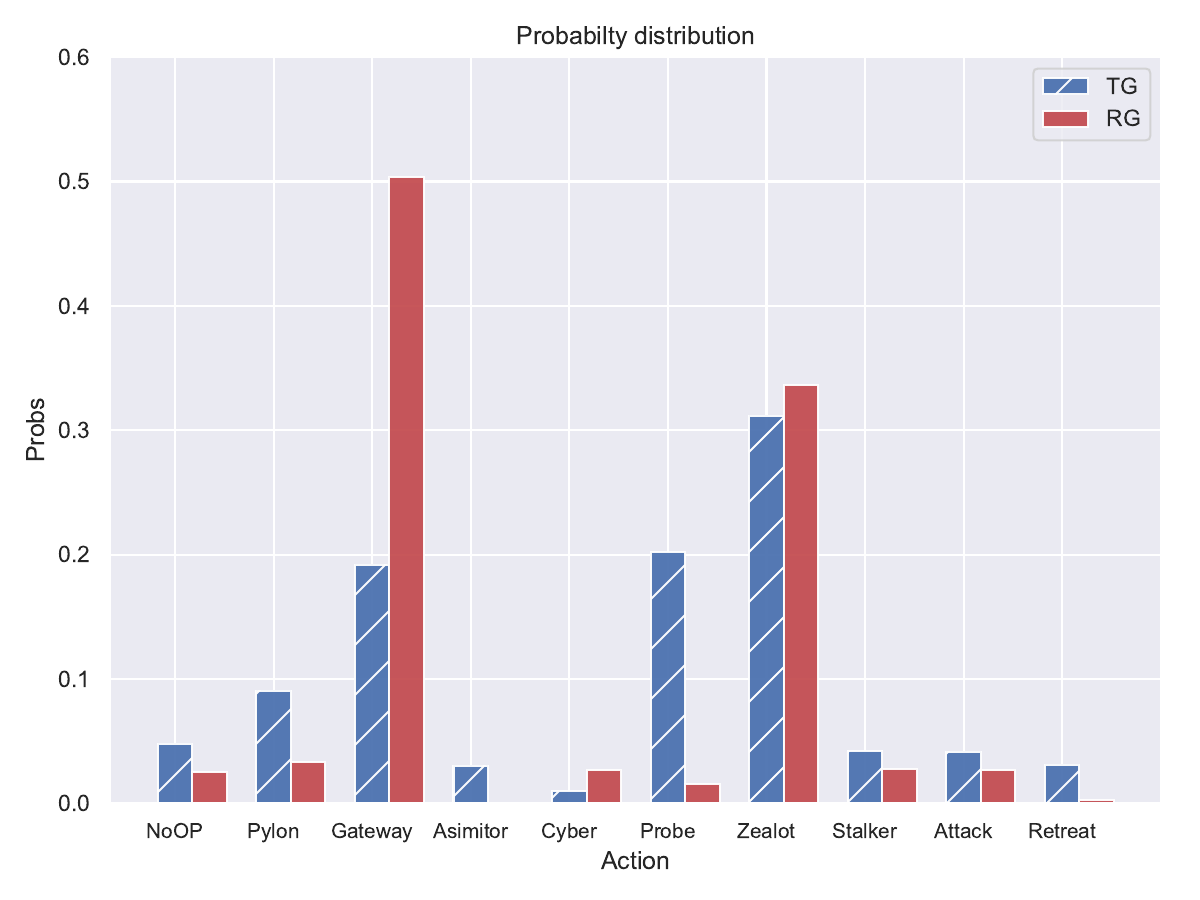}
        \label{fig:s.b.4.4.2}
    }
    \subfloat[]{
        \centering
        \includegraphics[width=0.475\columnwidth]{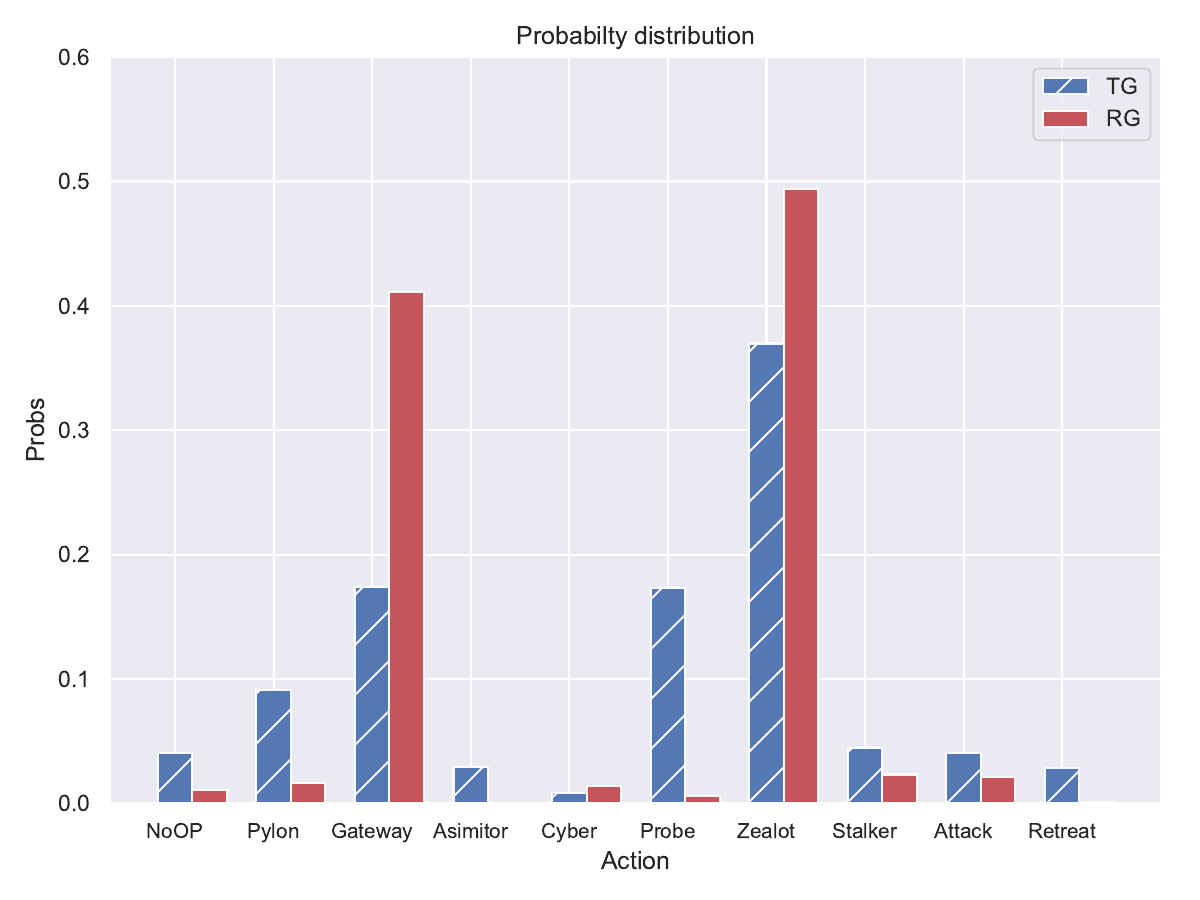}
        \label{fig:s.b.4.4.3}
    }
    \subfloat[]{
        \centering
        \includegraphics[width=0.475\columnwidth]{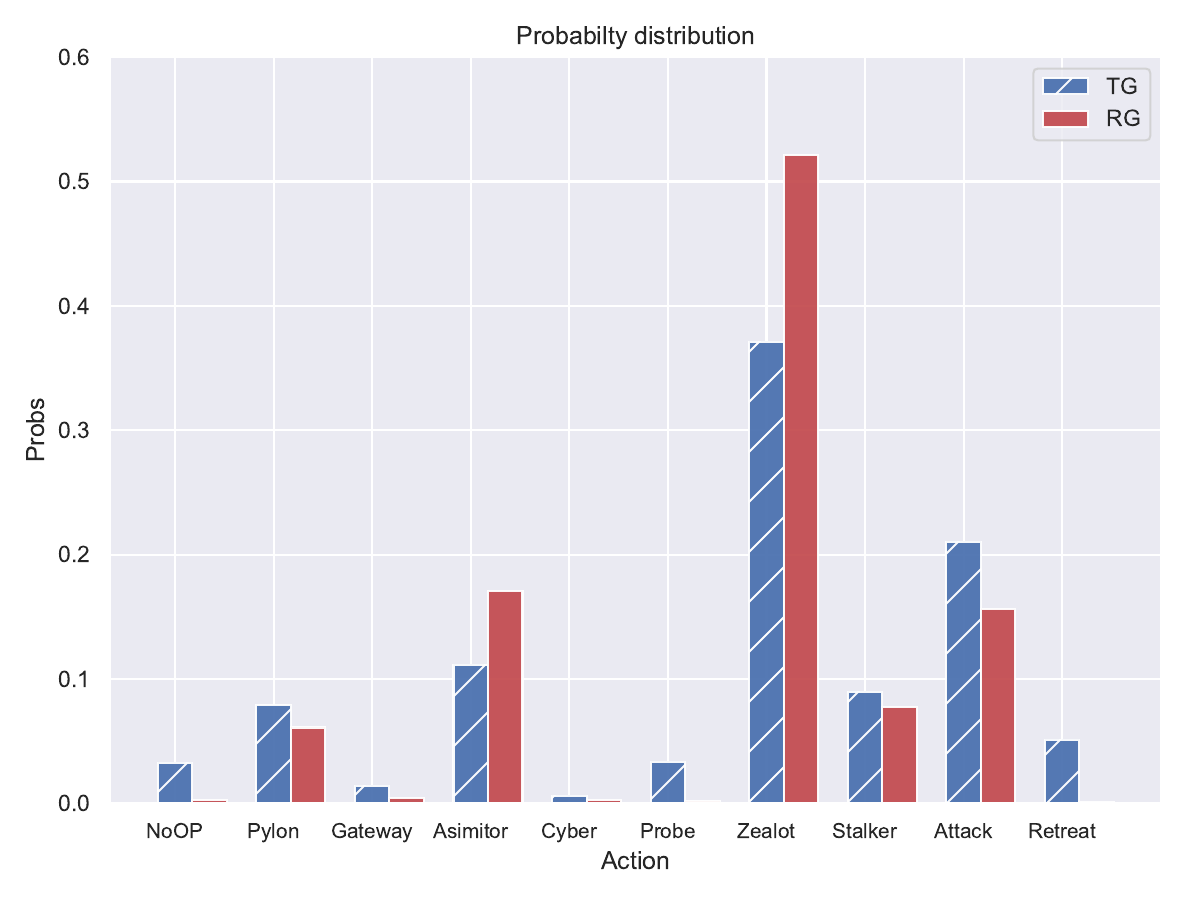}
        \label{fig:s.b.4.4.4}
    }%
    \caption{(a)~At 0s. (b)~At 30s. (c)~At 1min. (d)~At 3min. The action distribution of the agent at different time points in one episode after learning by $\varnothing \Rightarrow TG$ (shown in doted blue) or $TG \Rightarrow RG$ (shown in red).}
    \label{fig:s.b.4.4.a}
\end{figure*}
To give a better visualization, we show the action distributions of the trained agent at specific time points in one episode which reveals the difference between the trained policies in TG and RG (see Fig.~\ref{fig:s.b.4.4.a}). We can see that despite some actions has tendencies in the action distribution after being trained in TG, the distribution is still even. The transfer learning process in RG makes the agent concentrate on some actions which can defeat the L7 AI. 

In addition, we calculated the KL divergence of these two trained policies. We use $\text{KL}(TG \Rightarrow RG)$ to represent the KL divergence from the trained policy in RG to the trained one in TG. $\text{KL}(TG \Rightarrow RG)=0.53$, $\text{KL}(\varnothing \Rightarrow TG)=0.51$ and $\text{KL}(\varnothing \Rightarrow RG)=1.47$. Hence, 
 \begin{equation}
\text{KL}(\varnothing \Rightarrow TG) + \text{KL}(TG \Rightarrow RG) \ll \text{KL}(\varnothing \Rightarrow RG)
 \end{equation}
from this statistic perspective, process of decomposing improves the learning smoothness, due to transferring from one distribution to another one is easier.

\begin{table*}[t!]
\centering
    \scalebox{1.0}{
   \begin{tabular}{l c c c c c c c c c c}
       \toprule
       Method    &  Agent's Race  & O Race & Map & Level-1 & Level-2 & Level-3 & Level-4 & Level-5  & Level-6 & Level-7   \\    \midrule
       Pang's~\cite{pang2019sc2}    & Protoss & Terran & S64  & 100\% & 100\% & 99\%& 97\%& 100\%& 90\%& 93\%   \\
       Ours            & Protoss & Terran & S64  & 100\% & 100\%& \textbf{100}\%& \textbf{100}\%& \textbf{100}\%& \textbf{100}\%&  \textbf{99}\%  \\ \midrule
       Lee's~\cite{tang2018sc2}    & Zerg & Zerg & AR & \textbf{100}\% & \textbf{100}\% & \textbf{99}\%& 95\%& 94\%& 50\%& 60\% \\
       Ours             & Zerg & Zerg & AR & 98\% & 94\%& 97\%& \textbf{96}\%& \textbf{95}\%& \textbf{79}\%&  \textbf{69}\% \\    \midrule
       Race test 1         & Zerg & Zerg & S64 & 100\% & 100\% & 98\%& 97\%& 99\%& 96\%& 93\%\\
       Race test 2         & Terran & Terran & S64 & 100\% & 100\%& 97\%& 97\%& 95\%& 96\%&  95\%\\
       \midrule
       Map test 1            & Protoss & Terran & AR & 100\% & 100\% & 100\%& 98\%& 97\%& 96\%& 99\%\\
       Map test 2            & Protoss & Terran & F64 & 100\% & 100\%& 99\%& 99\%& 98\%& 98\%&  97\%\\
       \bottomrule
   \end{tabular}}
\caption{Performance results. We train our agent in level-7 and tested in the other six levels. O=Opponent's, AR=\emph{AbyssalReef}, S64=\emph{Simple64}, F64=\emph{Float64}. Each evaluation runs 100 games and is repeated for 3 times.}
\label{tab:evaluation results}
\end{table*}

% Pang's refer to \cite{pang2019sc2} and Lee's refer to \cite{tang2018sc2}

\begin{table}[h!]
    \centering
    \scalebox{1.0}{
    \begin{tabular}{l | c c c }
    \toprule
    Situations  & w/o & with \\
    \midrule
    $S_m < S_s, A_m = A_s$   & $0.01$ & $\textbf{0.96}$\\
    $S_m = S_s, A_m < A_s$  & $0.01$ & $\textbf{0.96}$\\
    $S_m < S_s, A_m < A_s$   &  $0.00$ & $\textbf{0.98}$\\
    $S_m < S_s, A_m < A_s, \exists S_{map}$  & $0.01$ & $\textbf{0.97}$\\
    \bottomrule
    \end{tabular}}
    \caption{Win-rate of using XfrNet or not while training 10000 episodes in $TG \Rightarrow RG$.}
    \label{tab:xfr net in 4 case}
\end{table}

\begin{table}[h!]
\centering
\scalebox{1.0}{
\begin{tabular}{l | c c c }
\toprule
Method  & Pang's~\cite{pang2019sc2} & Lee's~\cite{tang2018sc2} & Ours \\
\midrule
time (hours)   & $94.50$   & -     & $\textbf{1.08}$     \\
steps ( $10^6$ )      & $22.68$     & $6.00$     &  $\textbf{1.45}$   \\
\midrule
architecture  &   hierarchial          &     modular       &     \textbf{single}     \\
manual reward  &  yes        &    yes        &    \textbf{no}                            \\
\bottomrule
\end{tabular}}
\caption{Comparison of training time, training steps, architecture and reward design.}
\label{tab:training time}
\end{table}

\subsection{Comparison with other SC2 methods}
We compare our method with~\cite{pang2019sc2} and~\cite{tang2018sc2} which all used similar computational resources as us. The setting in~\cite{pang2019sc2} is as follows. They trained a Protoss agent and the opponent is a Terran bot. They test in the map \textit{Simple64}. Fig.~\ref{fig:s.a.1} (c) shows that the training process of our method is better than them. Note~\cite{pang2019sc2} is pre-trained on difficulty L2 and L5. The evaluation results are in Table~\ref{tab:evaluation results} which shows our win-rate is better than theirs. The setting in~\cite{tang2018sc2} is as follows. They trained a \textit{Zerg} agent, and the opponent is also a Zerg built-in AI. They test in the map \textit{AbyssalReef} from L1 to L7. We trained an agent using the same settings. Table~\ref{tab:evaluation results} shows that the performance of our method from L4 to L7 exceeds them. 

Our training time is also significantly better than the previous method (see Table~\ref{tab:training time}). When achieving the same win-rate as~\cite{pang2019sc2}, our training time is almost $1/100$ of theirs. We collect 3 run time samples of~\cite{pang2019sc2} and 5 ones of ours, then we found the comparison p-value is $0.0069$. Meanwhile, our training steps (consists of TG and RG) are also lower. Therefore, our approach is both superior in training speed and sample efficiency. Further, due to the effectiveness of TG, we use a simple neural network structure and no hand-designed reward. We didn't compare our method with methods by two companies that paying more than 100 times resources (from CPU cores or GPUs) than us. We concluded all RL methods based on pysc2 on the full game of SC2 in Table~\ref{tab:full comparison}.

\begin{table*}[h!]
    \centering
        \scalebox{1.0}{
        \begin{tabular}{l l l l l r r r l}
            \toprule
            Method  &Architecture &Methods &Knowledge &Micro action &CPU cores & GPU &Time (hours) &Performance \\
            \midrule
            Pang's~\cite{pang2019sc2}  &hierarchial &HRL &ER &None  &48 &4  &95  &Level-7\\           
            TStarBot~\cite{Sun2018tsbot} &hierarchial &HRL &MA &Use  &3840 &1 &24  &Level-10 \\           
            Lee's~\cite{tang2018sc2} &modular &SL, RL, SP &BO &Use &40 &8 &24  &Level-7\\          
            AlphaStar~\cite{AlphaStarNature} &modular &SL, RL, SP &ER, BO &Use &12000 &384 &1056 &GrandMaster\\         
            Ours &single &RL &BM &None &48 &4 &1 &Level-10\\
            \bottomrule
        \end{tabular}}
        \caption{Comparison of the RL methods on full-length SC2 by pysc2. SL=Supervised Learning, which is similar to behavior cloning. SP=self-play. ER=expert replays. BO=build orders. MA=hand-crafted macro actions. BM=builded model.}
        \label{tab:full comparison}
\end{table*}

\subsection{Comparision with WorldModel}
We compare our method with the WorldModel~\cite{Ha2018worldmodelNIPS} method on the full game of SC2. WorldModel is a more complicated model than~\cite{NagabandiKFL2018mbrl}. It uses VAE~\cite{VAE2014} and RNN~\cite{LSTM1997} to build a model with powerful representative capabilities. WorldModel performs well on the environment of \textit{CarRacing} and \textit{DOOM}. We compare our method with the two variants of the WorldModel which are \textit{world-model} and \textit{dream-model}. The world-model uses unsupervised learning to train the VAE and RNN on which a \textit{controller} is trained using evolution strategy (ES)~\cite{EvolutionStrategies2017} methods. It has no usage of the dynamic model. In contrast, the dream-model uses the trained VAE and RNN to build a dynamic model and then trains the controller on the dynamic model. Different from~\cite{Ha2018worldmodelNIPS}, we use RL instead of ES to train the controller (due to ES is slower on SC2). Fig.~\ref{fig:s.a.4.2.a}(c) shows the comparison in which the results of the world-model and dream-model are both not ideal. 

Reasons may be as follow: 1. The dimension compression characteristic of VAE may be more suitable for the natural images instead of the feature maps in SC2; 2. The exploration strength of random policy is limited, making it hard to capture the dynamic of the SC2 environment. In response to reason 2, we do an extended experiment. According to the description in the original paper, iteration training may solve the exploration problem. The new iteration training process is as follows: First, a dream model is trained, called dream-model-1; Second, we use policy trained by dream-model-1 to collect new trajectories on which a dream-model-2 is trained; Third, the previous process is repeated to train a dream-model-3. The results of iteration training are shown in Fig.~\ref{fig:s.a.4.2.a}(d). We can devise a conclusion that even the iteration training could not mitigate the exploration problem. Codes for our implementation of WorldModel-SC2 can be found in our codebase.

\begin{figure*}[h!]
    \centering
    \subfloat[]{
        \centering
        \includegraphics[width=0.475\columnwidth]{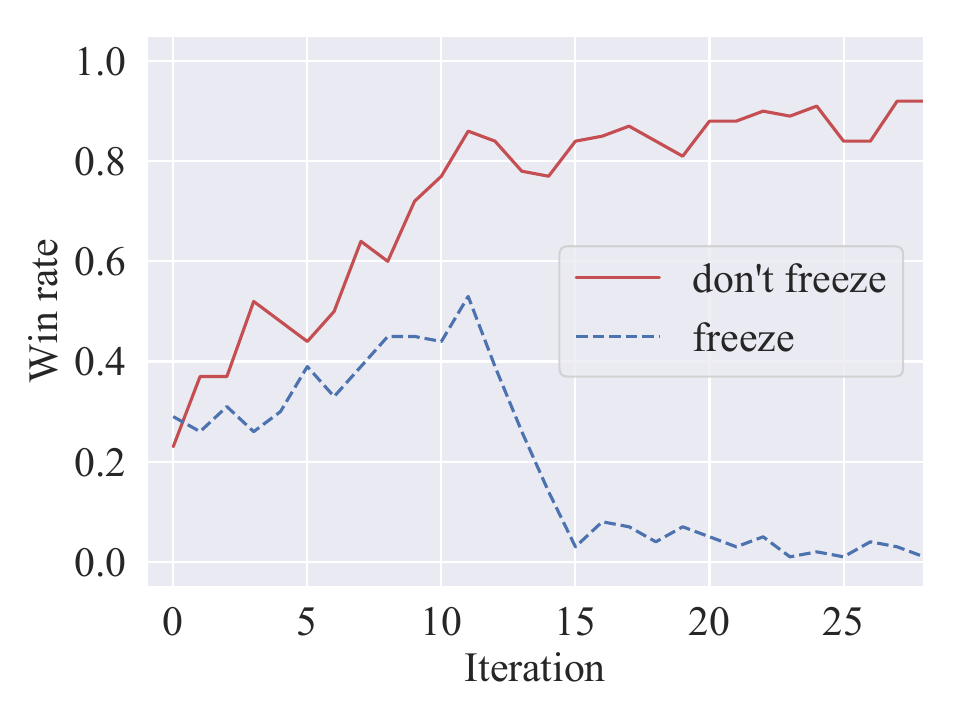}
        \label{fig:s.a.4.4.1}
   }
    \subfloat[]{
        \centering
        \includegraphics[width=0.475\columnwidth]{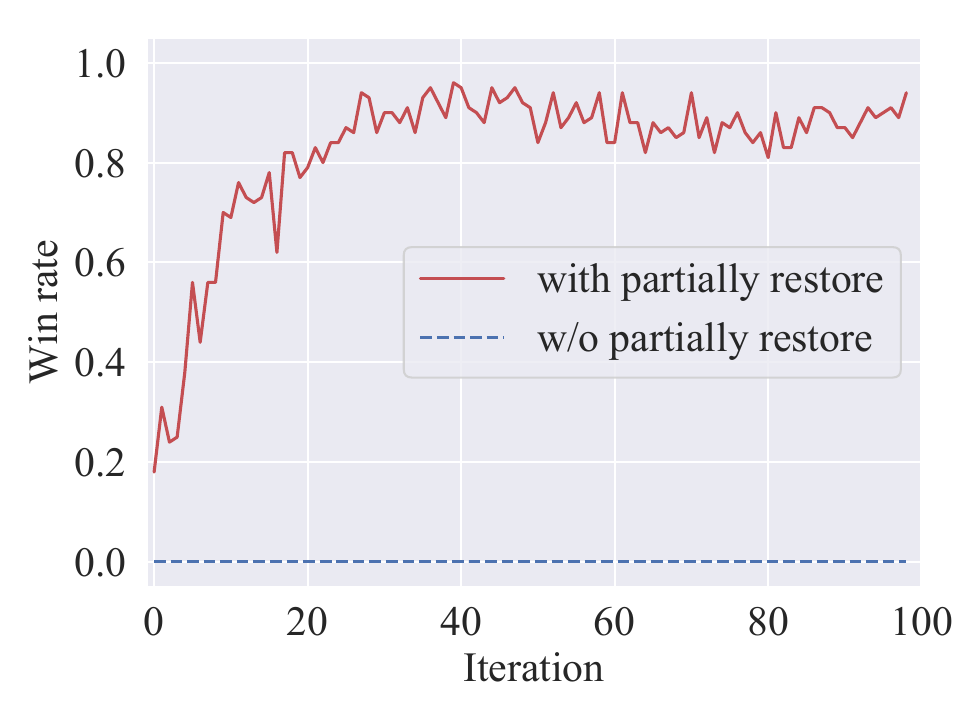}
        \label{fig:s.a.4.4.2}
    }
    \subfloat[]{
        \centering
        \includegraphics[width=0.475\columnwidth]{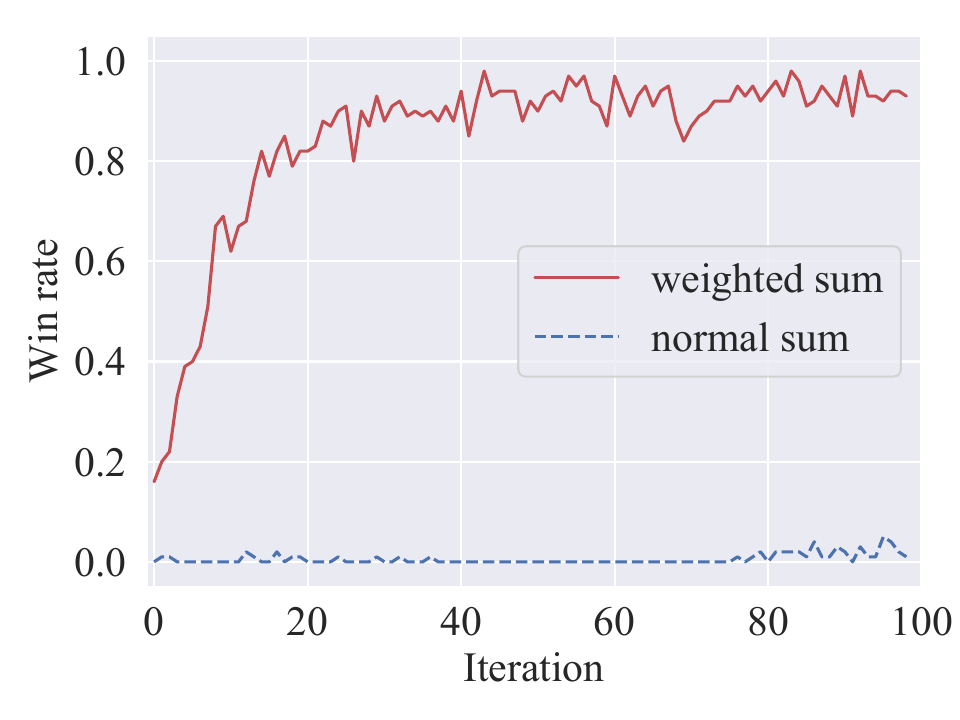}
        \label{fig:s.a.4.4.3}
    }
    \subfloat[]{
        \centering
        \includegraphics[width=0.475\columnwidth]{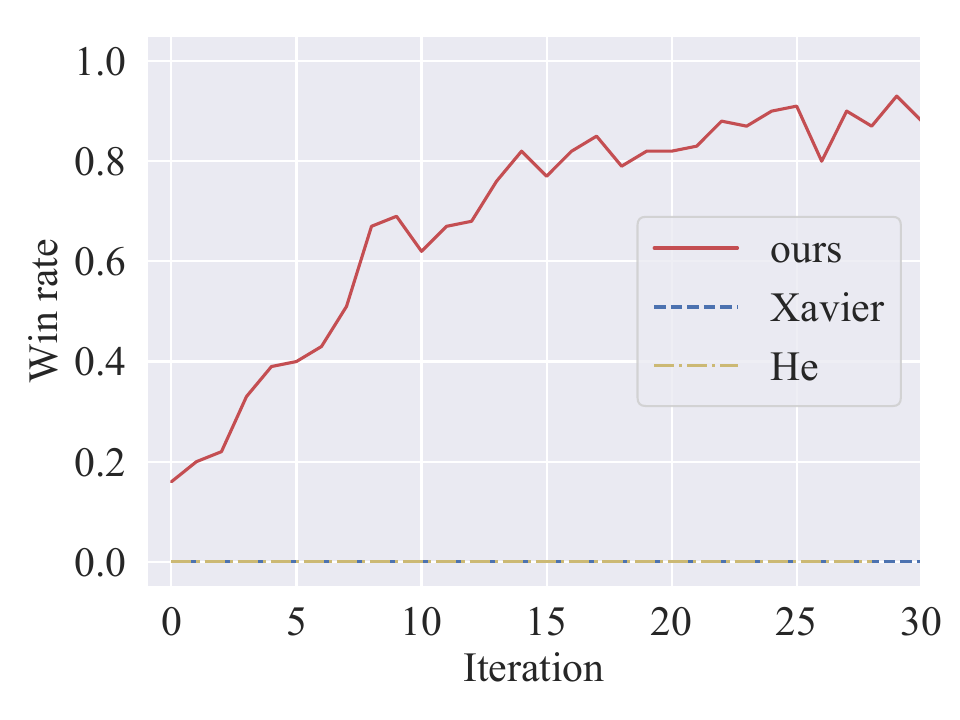}
        \label{fig:s.a.4.4.4}
    }%
    \caption{(a)~Using or using not freezing layers. (b)~Using or using not partially restore. (c)~Using or using not weighted sum. (d)~Effects of different initialization strategies. All experiments are done while $TG \Rightarrow RG $ with XfrNet.}
    \label{fig:s.a.4.4.a}
\end{figure*}

\subsection{Effect of XfrNet}
To validate the XfrNet's effectiveness, we make the following experiments. Table~\ref{tab:xfr net in 4 case} shows four different cases while $TG \Rightarrow RG$ (note that if we don't use XfrNet when state and action space are unequal, we must train from scratch). Moreover, we show the effects of using or not using the important mechanisms in the XfrNet, which is shown in Fig.~\ref{fig:s.a.4.4.a}. In addition, we found that when transferring with added actions, the probability of some actions among the newly added actions is already greater than the original actions', indicating that these new ones are indeed effectively trained. Combined with using XfrNet, by adding training in $RG \Rightarrow RG$ mode for fine-tuning in the high difficulty cheating-level and applying useful experiences which are presented in~\ref{section: tricks}, we can finally train an agent that beats the most difficult cheating L-10 built-in AI which is shown in Table~\ref{tab:against cheating}.

% Fig.~\ref{fig:s.b.4.4.d}

\begin{table}[h!]
    \centering
    \scalebox{1.0}{
    \begin{tabular}{l | c c c }
    \toprule
    Difficult  & Level-8 & Level-9 & Level-10 \\
    \midrule
    Pang's~\cite{pang2019sc2}  &   0.74   &  0.71   &  0.43 \\
    Ours  &   \textbf{0.95}   &  \textbf{0.94}   &  \textbf{0.90} \\
    \bottomrule
    \end{tabular}
    }
    \caption{The average win-rate against cheating-level AIs.}
    \label{tab:against cheating}
\end{table}

%\subsection{Thought-Game Hypothesis}
\subsection{Influence of fidelity levels of TG}
In order to verify the TG hypothesis, we did experiments on SC2. The expert divides the rules of RG into $4$ parts (see Fig.~\ref{fig:modules}) each of which has several modules. By the importance of these modules, we can think the order of importance of these parts is: Combat, Logistics, Production, and Resource. We then designed different fidelity levels of TG which are defined as below (note the first equation means TG$^3$ only abstracts the modules in the combat part): $\text{TG}^3=\{\text{Combat}\}, \text{TG}^2=\text{TG}^3+\{\text{Logistics}\}, \text{TG}^1=\text{TG}^2+\{\text{Production}\}, \text{TG}^0=\text{TG}^1+\{\text{Resource}\}$. If we abstract more rules from TG to RG, the difficulty of learning in TG increase (see Fig.~\ref{fig:s.a.1} (a)), but the difficulty of learning in RG decreases (see Fig.~\ref{fig:s.a.1} (b)), which verifies the TG hypothesis.

\begin{figure*}[h!]
    \centering
    \subfloat[]{
        \centering
        \includegraphics[width=0.475\columnwidth]{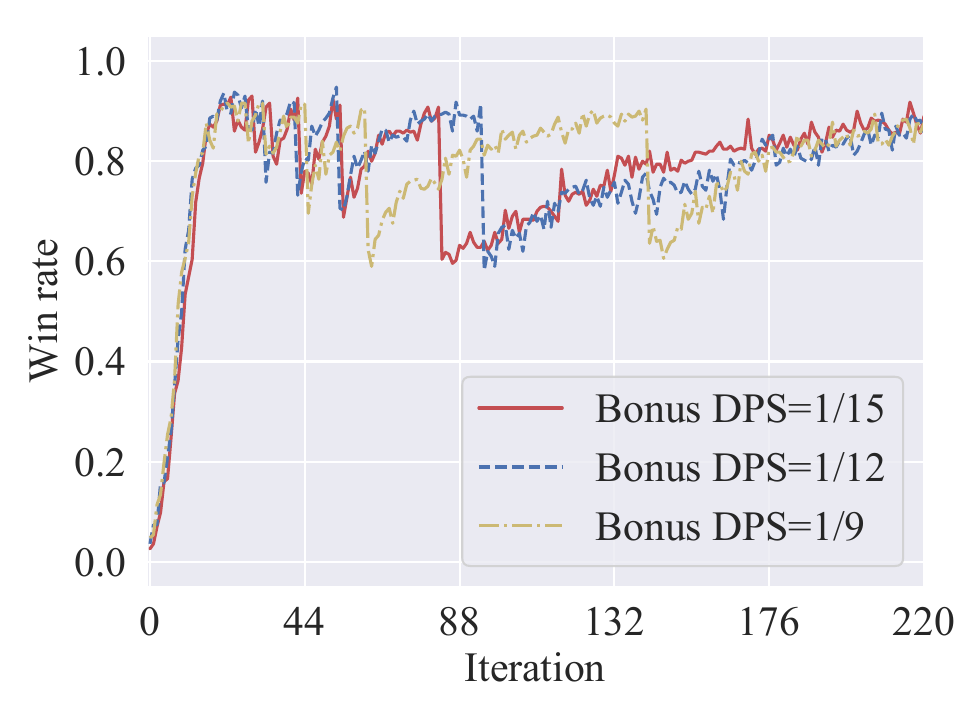}
        \label{fig:s.2.7.2}
   }
    \subfloat[]{
        \centering
        \includegraphics[width=0.475\columnwidth]{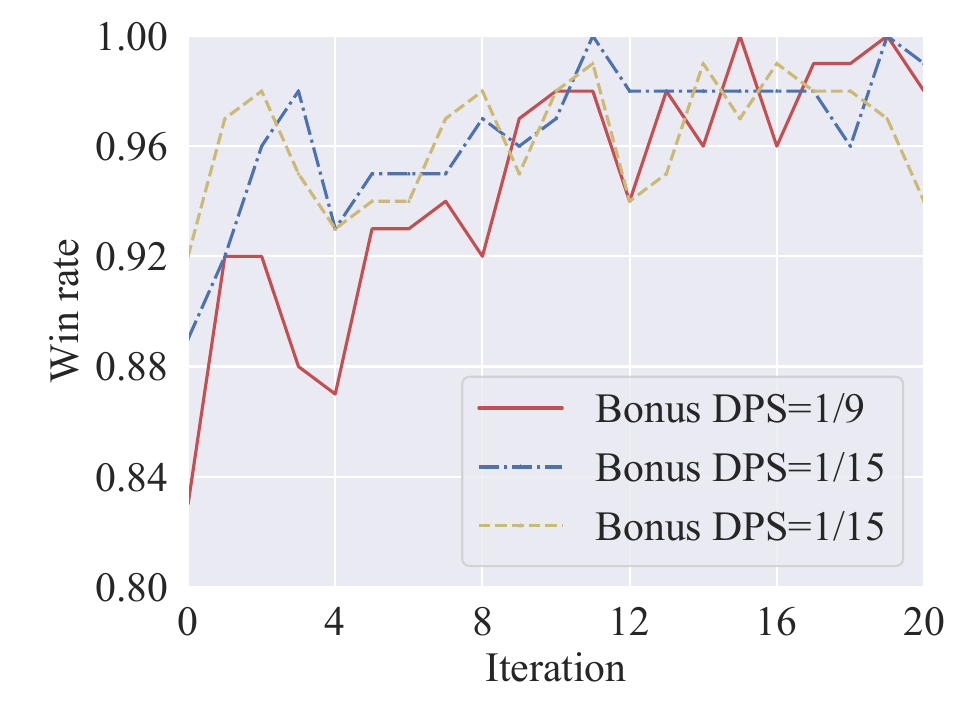}
        \label{fig:s.2.8.1}
    }
    \subfloat[]{
        \centering
        \includegraphics[width=0.475\columnwidth]{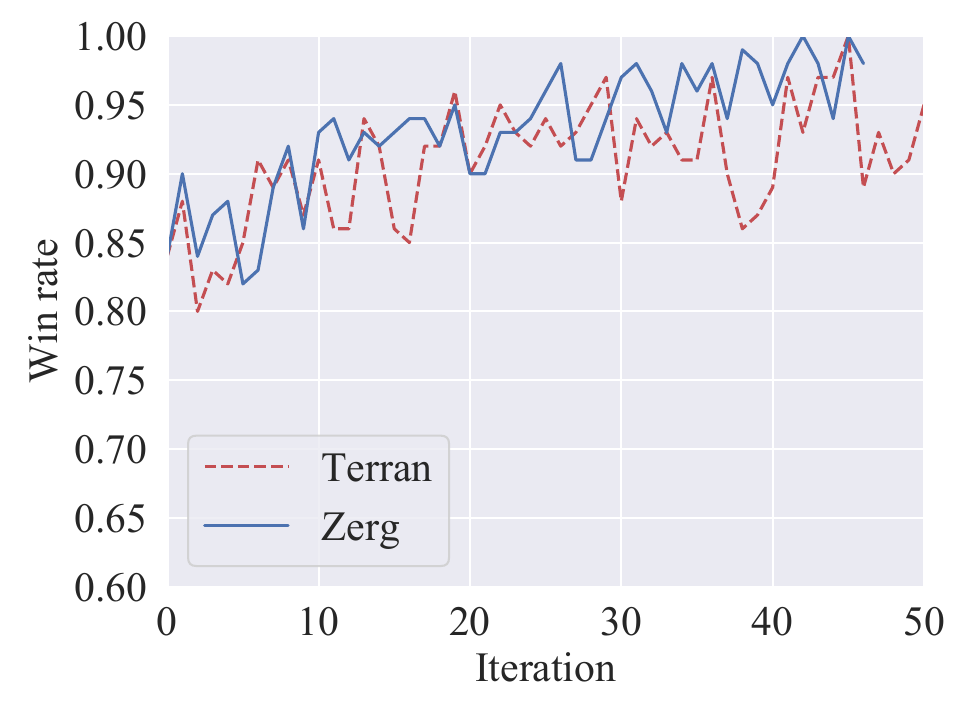}
        \label{fig:s.2.9.2}
    }
    \subfloat[]{
        \centering
        \includegraphics[width=0.475\columnwidth]{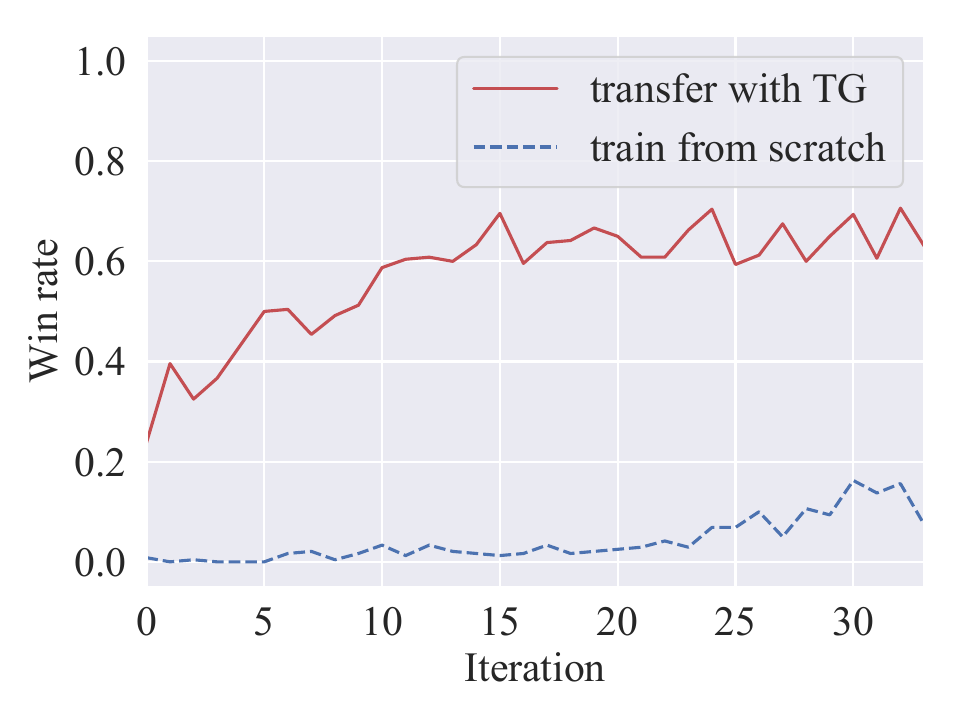}
        \label{fig:s.2.0.1}
    }
    \caption{(a)~Changing the bonus DPS while $\varnothing \Rightarrow TG$. (b)~Changing the bonus DPS while $TG \Rightarrow RG$.
    (c)~Training of \textit{Zerg} and \textit{Terran} while $TG \Rightarrow RG$. (d)~$TG \Rightarrow RG$ vs. $\varnothing \Rightarrow RG$ on SC1.}
    \label{fig:s.a.2}
\end{figure*}

\begin{figure*}[h!]
    \centering
    \subfloat[]{
        \centering
        \includegraphics[width=0.475\columnwidth]{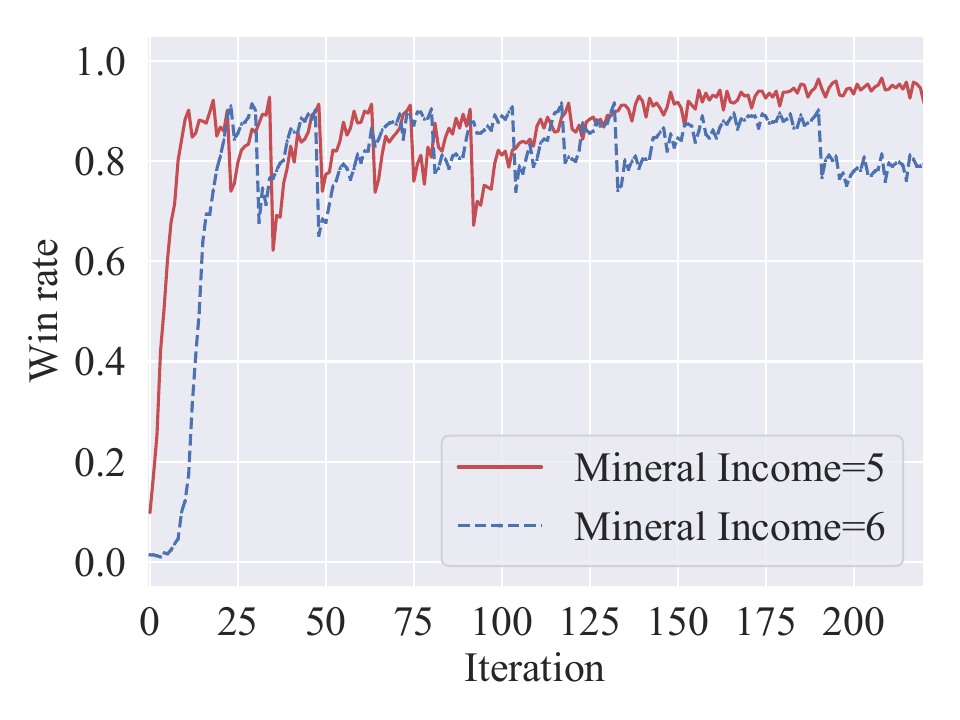}
        \label{fig:s.2.7.2}
   }
    \subfloat[]{
        \centering
        \includegraphics[width=0.475\columnwidth]{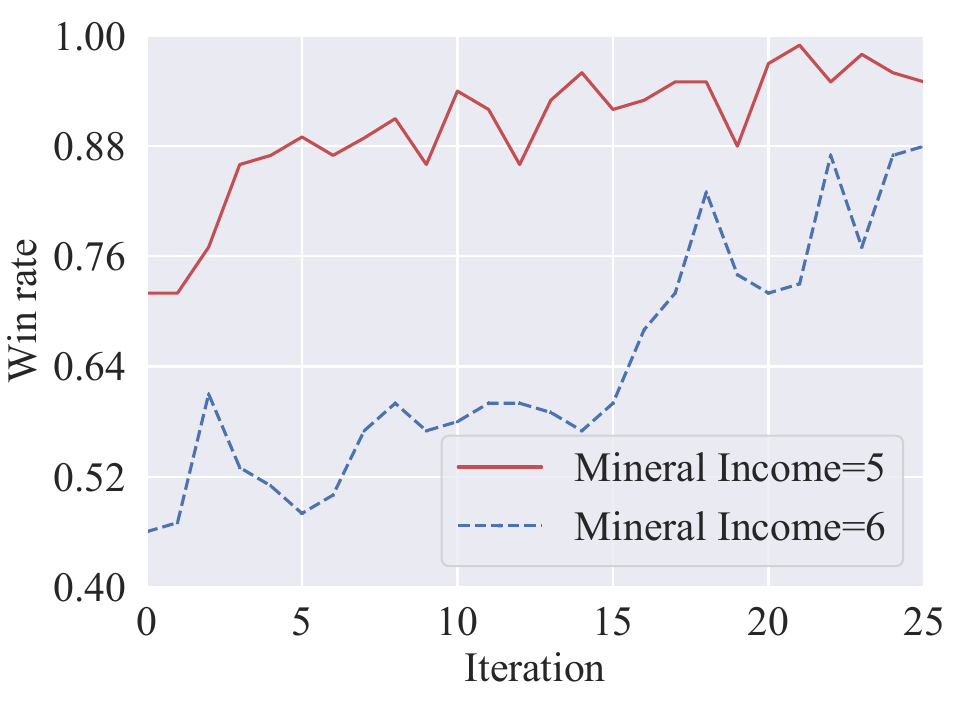}
        \label{fig:s.2.5.1}
    }
    \subfloat[]{
        \centering
        \includegraphics[width=0.475\columnwidth]{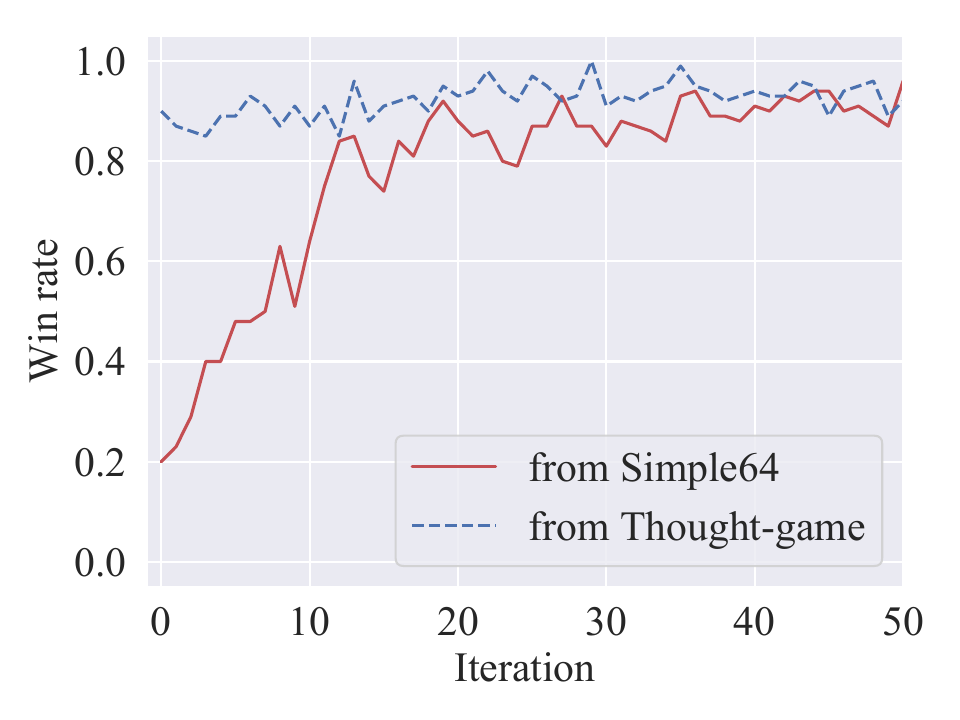}
        \label{fig:s.4.8.2}
    }%
    \subfloat[]{
        \centering
        \includegraphics[width=0.475\columnwidth]{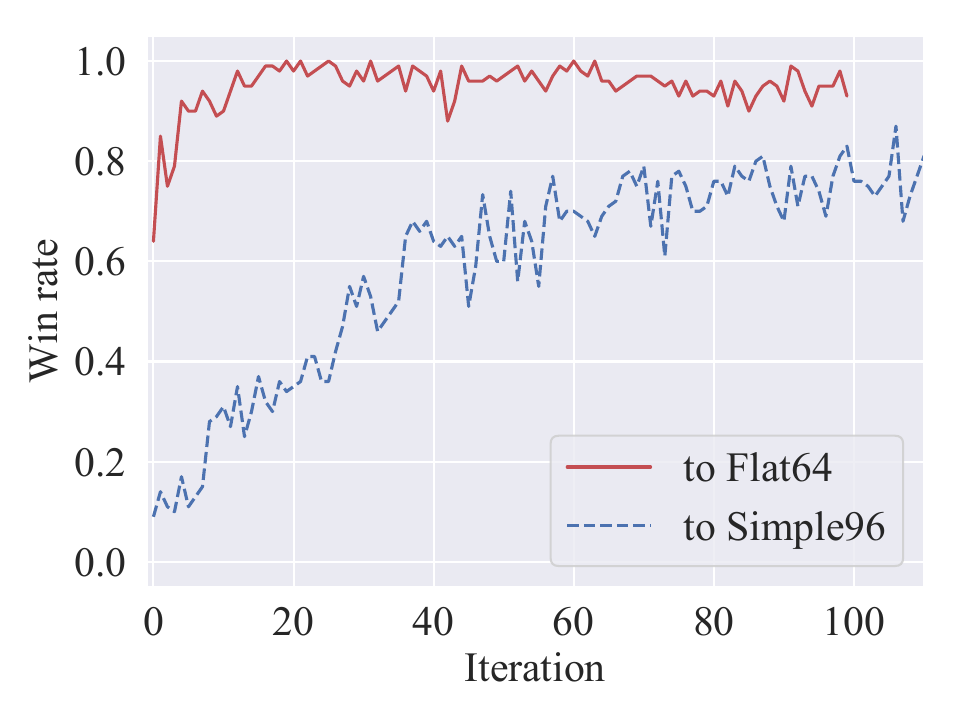}
        \label{fig:s.4.8.3}
    }%
    \caption{(a)~Changing economy parameter while $\varnothing \Rightarrow TG$. (b)~Changing economy parameter while $TG \Rightarrow RG$. (c)~Using different initial policy to transfer to AbyssalReef. (d)~Using the same initial policy to transfer to different maps.}
    \label{fig:s.a.3}
\end{figure*}

%\subsection{Impact of Parameters in Thought-Game}
\subsection{Impact of rule-parameters in TG}
We test the effect of making changes to the rule-parameters of TG. We modify the two important rule-parameters which affect combat (DPS, damage per second) and economy (mineral income per one worker and one step). As can be seen from Fig.~\ref{fig:s.a.2} (a) and Fig.~\ref{fig:s.a.2} (b), the impact of combat rule-parameters is small (economy-parameter results on TG and RG are in Fig.~\ref{fig:s.a.3} (a) and Fig.~\ref{fig:s.a.3} (b)). In fact, this conclusion is also in line with our assumptions. i.e., using TG, the agent learns the rules in the RG, not the specific rule-parameters.

%\subsection{Ablation Experiment}
\subsection{Difference between using TG and not}
In order to provide an ablation experiment on the effectiveness of TG, we have given a baseline, which is the effect of training without TG. The result is shown in Fig.~\ref{fig:s.a.1} (d), which compares the effects of using the TG for pre-training with training from scratch. The run of training from scratch has been trained for 5 iterations (about 10 minutes, to simulate the training time in TG to make comparison fair). Each evaluation runs 3 times and the results are averaged.

\subsection{Analysis of performance and time}
Here we analyze why our agents get better performance. We argue this may be due to the characteristics of DL and RL. A better initialization will result in a better effect in DL. Meanwhile, a good initial policy which gives more reward signal at first may also make RL learning faster. 

The consumption of training time consists of several parts: 1. The time $t_\omega$ sampling in the environment, which depends on the simulation speed of the environment. 2. The required sample size $m_\mu$ of the training algorithm. The total sample size of the training is the sum of the sample sizes on all tasks in the context of curriculum learning. 3. The time cost of the optimization algorithm, $t_\eta$. Therefore, the total time overhead is $T_1 = (t_\omega+t_\eta) m_\mu$. Sampling time is usually longer than optimization time, \emph{i.e.}, $t_\omega \gg t_\eta$, hence $T_1 \approx t_\omega m_\mu$. Suppose our training process divides the task into $k$ steps. Step $k$ is our target task, and the previous step $k-1$ is the pre-training process for curriculum learning. Therefore, we can denote our training time as $t_\omega  m_\mu = t_\omega  (m_{\mu_1} + m_{\mu_2} + \cdots + m_{\mu_k})$. In our approach, we moved the part of the curriculum to the TG. Assume the time $t_\xi=1/M \cdot t_{\omega}$ in the TG, and the amount of sample required by the last step of the task $ m_{\mu_k}$ is $1/K$ of the total $m_\mu$. Then the time $T_2$ required by our algorithm can be written as $T_2 = t_{\xi} (m_{\mu_1}+m_{\mu_2}+...+m_{\mu_{k-1}})+ t_{\omega} m_{\mu_k} \approx 1/M \cdot t_{\omega}  m_\mu + 1/K \cdot t_{\omega} m_\mu = (1/M+1/K)T_1$. It shows that the speed-up ratio of the new algorithm is determined by the smaller value of the acceleration ratio of $M$ and the last task sample ratio $K$. So speed boost comes from the increase of simulation speed (due to simplicity of TG) and the migration of the curriculum from RG to TG (by using ACRL).

%\subsection{Expansibility Experiments}
\subsection{Results of applying TG to other situations}
There are three races in SC2 of which the units and buildings are quite different which means that their state and action space are much different. In order to test the extensibility of TG, we hope to test in the other two races than Protoss, which are Zerg and Terran. This means we need to design two different agents  (as AS does) and the corresponding TG. We use the GTG process to build the TG-Zerg and TG-Terran. The agents we finally trained can beat the L7 built-in AI (in Table~\ref{tab:evaluation results}). We also trained our agent on other maps of SC2 to show the applicability of our method of which the processes can be seen in Fig.~\ref{fig:s.a.3} (c) and Fig.~\ref{fig:s.a.3} (d) and the results are in Table~\ref{tab:evaluation results}. 

Can the TG method be applied to another game? We selected SC1 for testing. Differences exist between SC1 and SC2, e.g., their state and action spaces are different. We use the Protoss as race and train the agent against a Terran built-in bot on the map \textit{Lost Temple} by using the simulator \textit{torchcraft}. We generated a TG-SC1 and used the TTG algorithm to train the agent. We find the performance of our agent surpassed the effect of not using TG which is shown in Fig.~\ref{fig:s.a.2} (d).

\begin{figure*}[h!]
    \centering
    \subfloat[]{
        \centering
        \includegraphics[width=0.475\columnwidth]{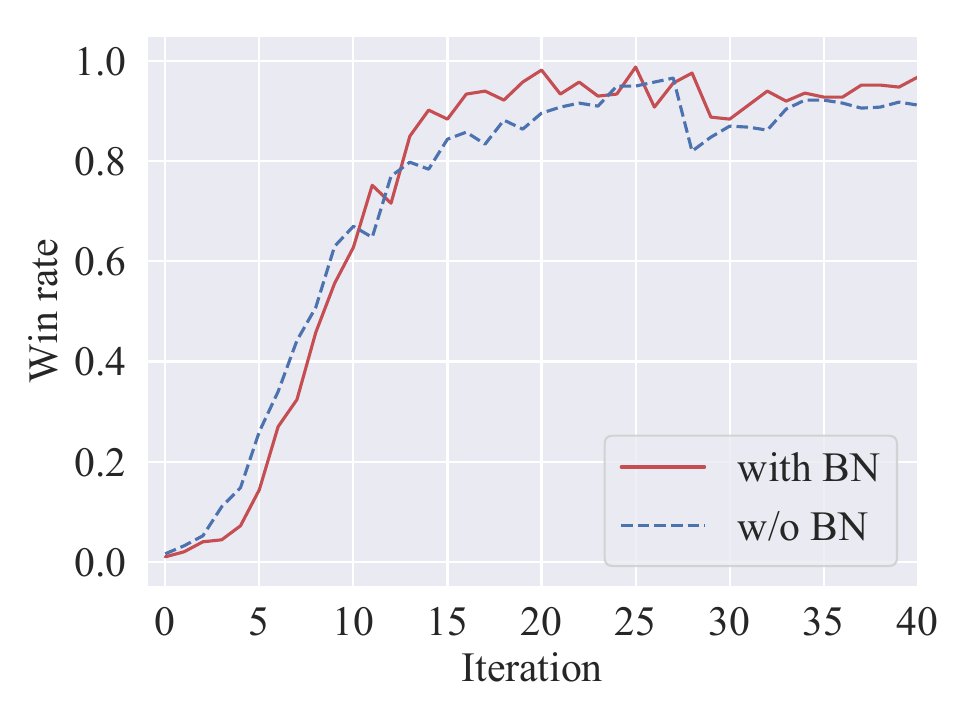}
        \label{fig:s.a.4.3.6}
   }
    \subfloat[]{
        \centering
        \includegraphics[width=0.475\columnwidth]{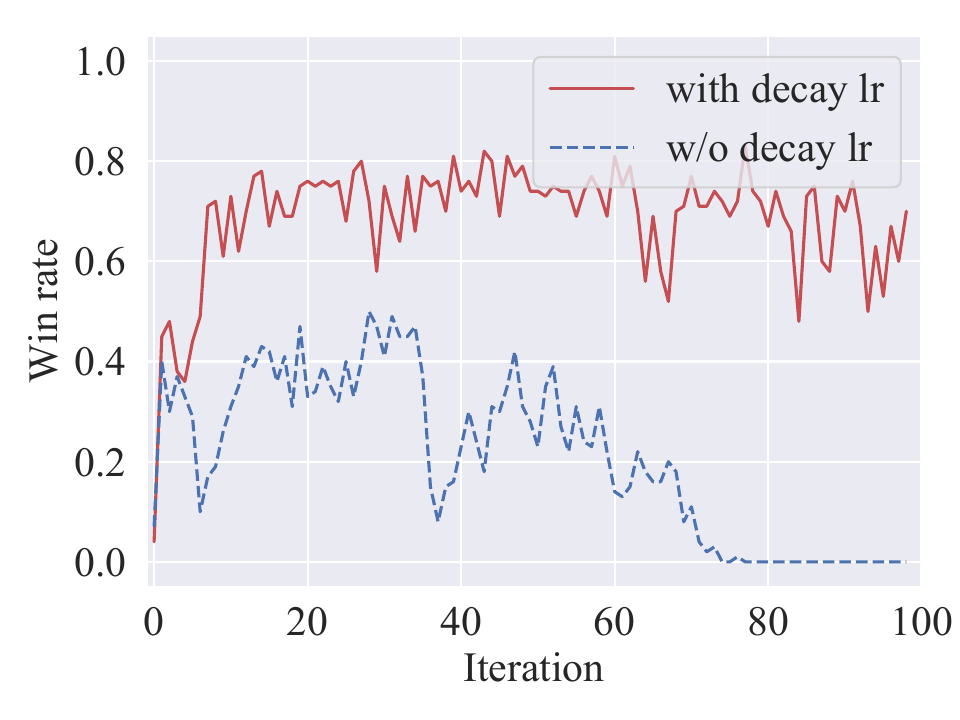}
        \label{fig:s.a.4.3.5}
    }
    \subfloat[]{
        \centering
        \includegraphics[width=0.475\columnwidth]{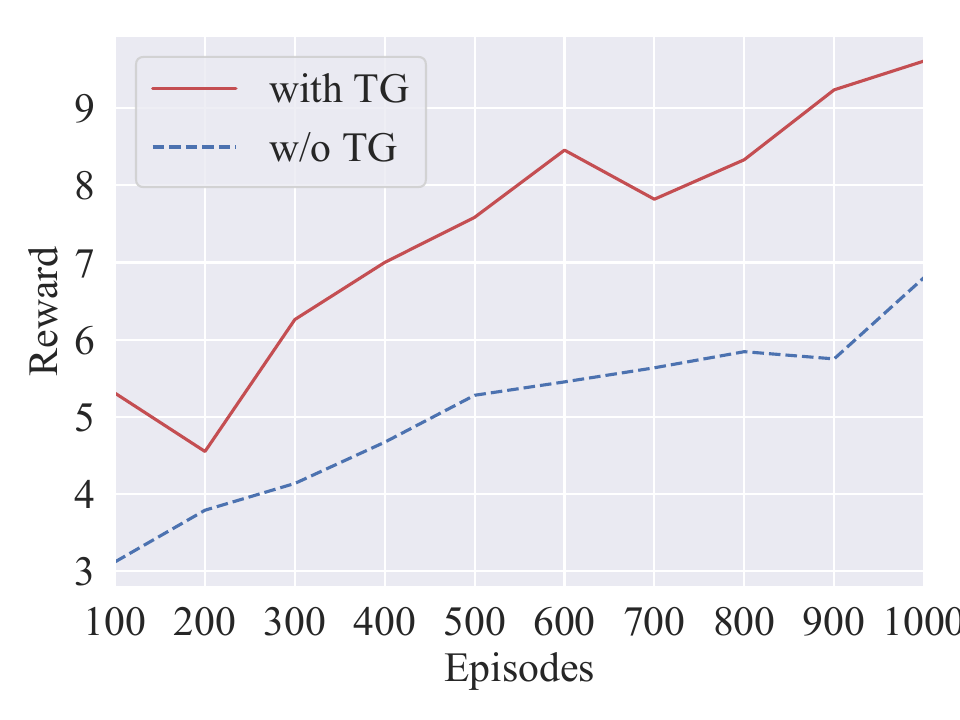}
        \label{fig:s.a.4.2.6}
    }
    \subfloat[]{
        \centering
        \includegraphics[width=0.475\columnwidth]{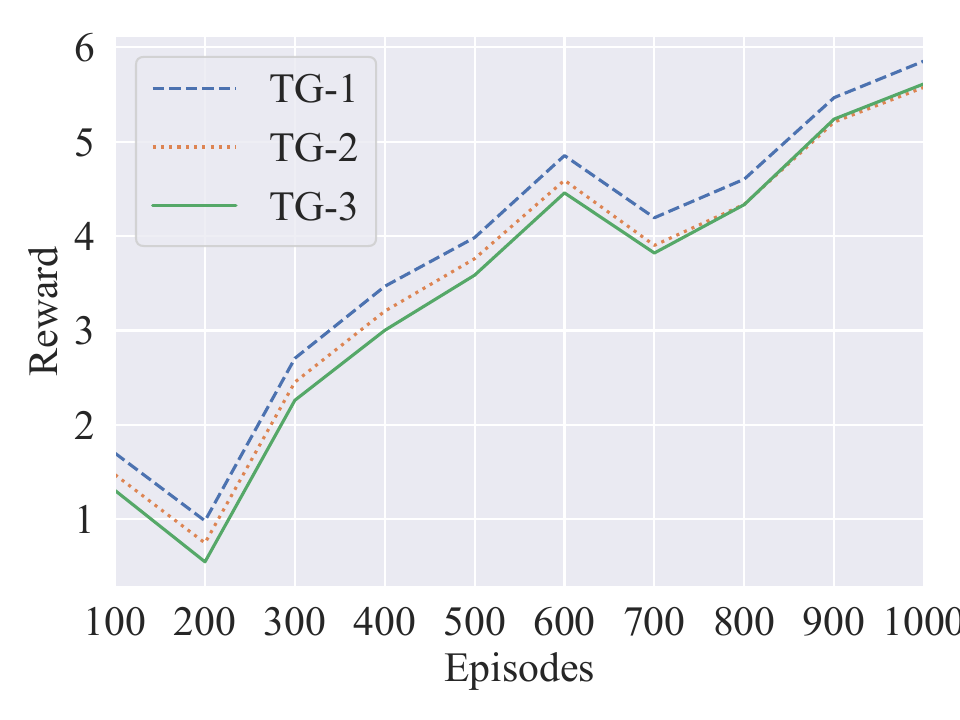}
        \label{fig:s.a.4.2.5}
    }%
    \caption{(a)~Effects of BN while $\varnothing \Rightarrow TG$. (b)~Effects of learning rate decay strategy with BN while $TG \Rightarrow RG$. (c)~Using TG in a hydropower task. (d)~TG hypothesis in the hydropower task.}
    \label{fig:s.a.4.2.c}
\end{figure*}

\subsection{Results of applying TG to real-world tasks}
SC1 shares some similarities with SC2. Can we apply TG in a completely different environment? To answer it, we evaluate TG on a real-world task which is a hydropower task needing control of a real water reservoir located in China Sichuan province to maximize its generation power in a fixed period of time. We build the TG based on the simulator of the reservoir. The simulator contains many real rule-parameters of the reservoir while our TG uses unreal ones. E.g., calculation the power ($P$) when transiting to the next state is influenced by \textit{water\_head} ($W$), \textit{generator\_release} ($G$), and \textit{generator\_params}. Normally, the more $G$, the less $W$. The generator\_params consist of many real rule-parameters of the reservoir and these parameters will participate in the calculation of power, making it a complex formula. But, it follows the monotonical trend that $P \propto W$ and $P \propto G$. Hence, our abstract rule is $P=\alpha W G$ where $\alpha$ is a hyper-parameter (we set it to $1\text{e-}5$ to prevent overflow). We found the policy learned with TG converges faster and has a better result than without it (see Fig.~\ref{fig:s.a.4.2.c}(c)). Then, we divided the TG into three different fidelity levels and found that it also meets the previously proposed TG hypothesis, as shown in Fig.~\ref{fig:s.a.4.2.c}(d). Note that we calculate the average episode reward of the last 100 episodes every 100 episodes and use the same random seed for all runs.

% We give the general abstract discipline instead of putting any real rule-parameters in the calculation. We found that even if the policy learned by the agent in TG is not accurate, it is still in the right direction, so the learning speed of transferring to RG is much faster than training from random initialization. 

\subsection{Effect of applying TG to play against humans}
For testing the performance of our method, we also test our agent against two human players. The two human players are an SC2 novice and an SC2 \textit{Golden} level player. Human players are restricted to not choosing blocking tactics (blocking tactics mean, e.g., the player uses some buildings to block the entrance of his main-base) because the agent did not see any opponents using blocking tactics at training time (before the game, human players knew these restrictions and agreed) \footnote{One video of the replays can be found at \url{https://drive.google.com/file/d/1RVFpJIhSvR0HL8dtEkjtmM6OkSMRy1NQ/view?usp=sharing}}. Human players can use a series of micro while our agent uses only macro-actions. Due to our agent make dicisions every two seconds, making its APM (action per minute) not larger than humans. The results are in Table~\ref{tab:fight with human}. Please note that at the time playing against humans, we use the agent which beats L-7 built-in AI, not the one which can beat L-10 AI.

% https://drive.google.com/file/d/1RVFpJIhSvR0HL8dtEkjtmM6OkSMRy1NQ/view?usp=sharing
% https://drive.google.com/file/d/1EucWio3YMUmXv3dvXEL-RUhh-jr2XSPt/view?usp=sharing

\begin{table}[h!]
\centering
\scalebox{1.0}{
\begin{tabular}{l | c c c | c  }
\toprule
Player  &  A Race  & H Race & Map & Result \\
\midrule
SC2 novice      & Protoss & Terran & S64    & 5:0     \\
SC2 golden      & Protoss & Terran & S64    & 4:1     \\
\bottomrule
\end{tabular}
}
\caption{Play against human. A=Agent, H=Human, S64=\emph{Simple64}. Result=(agent:human)}
\label{tab:fight with human}
\end{table}

\begin{figure*}[h!]
    \centering
    \subfloat[]{
        \centering
        \includegraphics[width=0.475\columnwidth]{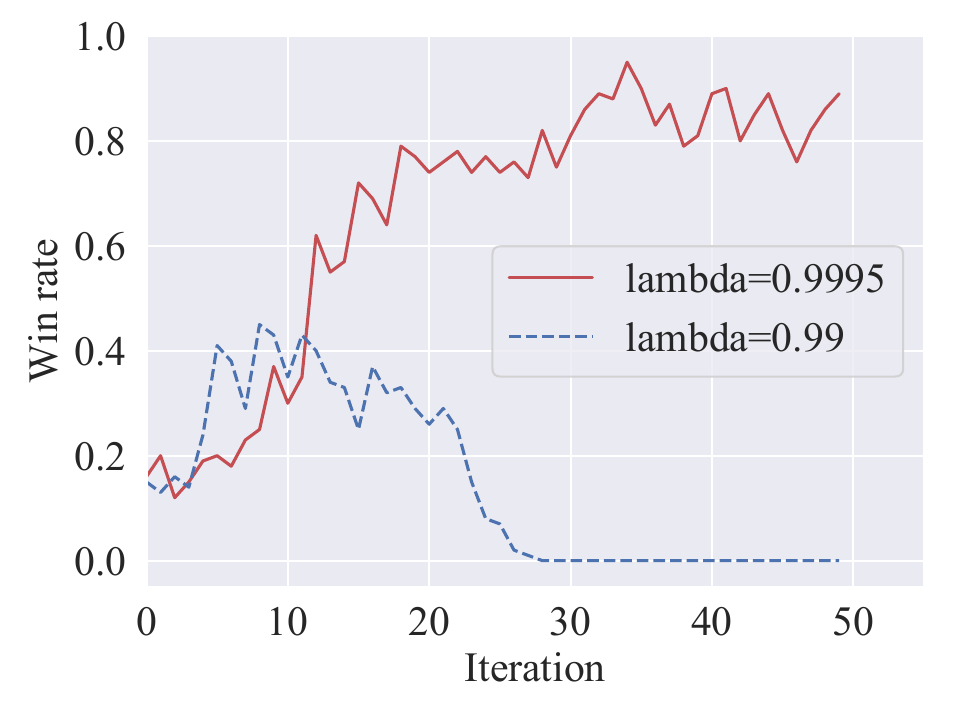}
        \label{fig:s.a.4.3.1}
    }
    % \hspace{0.01cm}
    \subfloat[]{
        \centering
        \includegraphics[width=0.475\columnwidth]{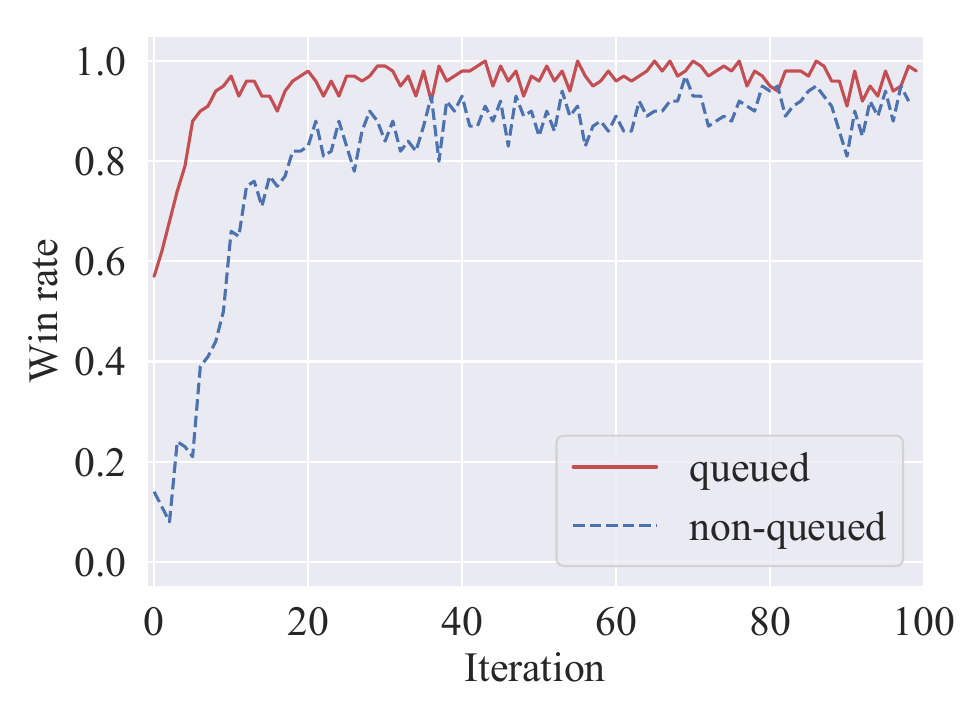}
        \label{fig:s.a.4.3.2}
    }
    % \hspace{0.01cm}
    \subfloat[]{
        \centering
        \includegraphics[width=0.475\columnwidth]{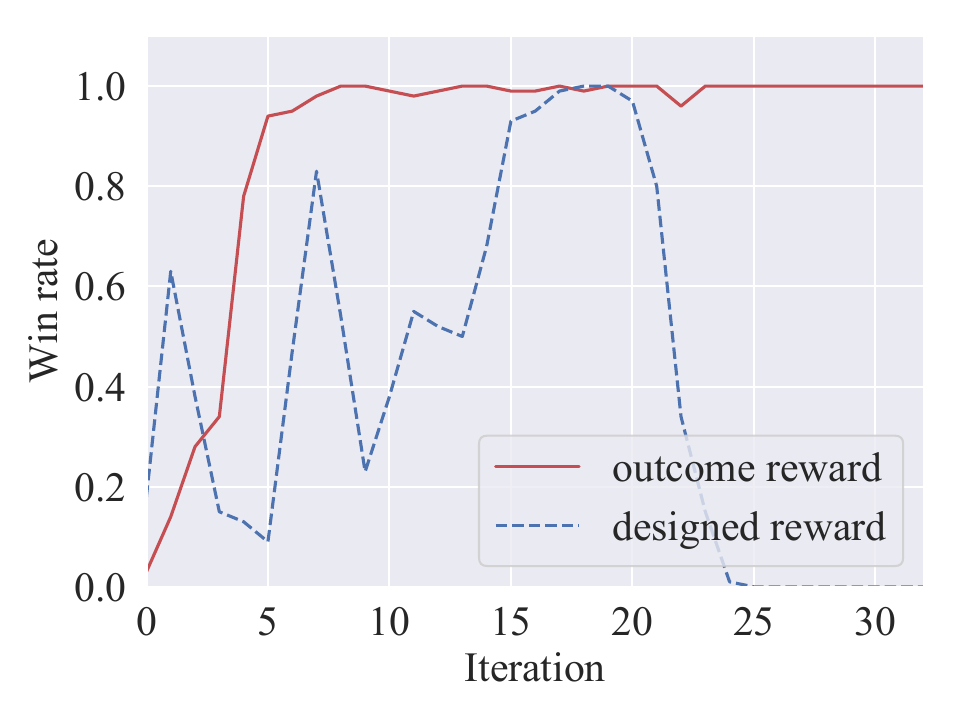}
        \label{fig:s.a.4.3.3}
    }
    % \hspace{0.01cm}
    \subfloat[]{
        \centering
        \includegraphics[width=0.475\columnwidth]{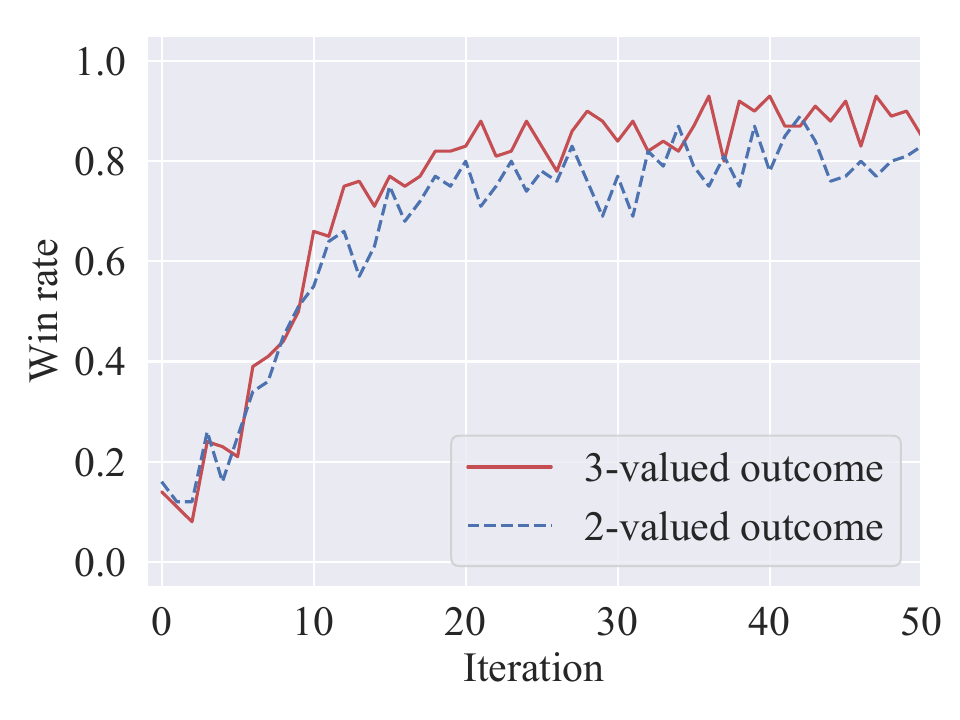}
        \label{fig:s.a.4.3.4}
    }
    \caption{(a)~Different $\lambda$ of GAE. (b)~Queued vs. non-queued actions. (c)~Outcome reward vs. designed reward. (d)~2-valued vs. 3-valued outcome reward. All experiments are done while $TG \Rightarrow RG$.}
    \label{fig:s.a.4.3.a}
\end{figure*}

\subsection{Important hyper-parameters on training} \label{section: tricks}
Through experiments, we found some hyper-parameters have higher effects on training:

1. The hyper-parameter $\lambda$ in GAE. $\lambda$ is often be defaulted to set to 0.99. We found that this setting may bring deterioration in the learning process which is shown in Fig.~\ref{fig:s.a.4.3.a}(a). Through analysis, we found that when the outcome reward is 1, the advantage estimator passing to the starting state will become very small, e.g, it be $0.99^{300} \approx 0.049$ (one episode often lasts for 300 steps in our setting on SC2). Due to that, we modify the lambda to $0.9995$, making the advantage estimator at the start increase to $0.9995^{300} \approx 0.86$. After this change, we found the stableness and speed of learning increase and the phenomenon of deterioration happens rarely.

2. Queued action. One can give the actions of SC2 a boolean argument called \textit{queue}. If it is true, the entity executes the commanded action after it has executed the action at hand. If it is false, the entity will cancel the action at hand and execute the commanded action. The queue argument affects the learning which is shown in Fig.~\ref{fig:s.a.4.3.a}(b).

3. If we use a manual reward design, but the design is not good, it will cause the following problem: the win-rate grows fast at first dozens of iterations, but then fall down quickly. This unstable training is shown in Fig.~\ref{fig:s.a.4.3.a}(c). Hence we recommend using outcome reward as the training signal. 

4. To increase efficiency, we set a max step for every episode in SC2. So, when one episode is over, the game may not have a true winner. Thus we may have a 2-valued $\{0, 1\}$ (means not-win and win) or 3-valued $\{-1, 0, 1\}$ outcome. We found 3-valued is better because it encourages the agent to defend its home to gain a draw when at the start of training which is shown in Fig.~\ref{fig:s.a.4.3.a}(d). 

5. Batch normalization (BN)~\cite{BatchNormalization2015} is a widely used normalization strategy in DL. We found that we can train the agent with BN while $\varnothing \Rightarrow TG$ which is shown in Fig.~\ref{fig:s.a.4.2.c}(a). However, when training in $TG \Rightarrow RG$, due to the change of the range of state features, the hyper-parameters of BN layer $\gamma$ and $\beta$ need to have a big adjustment. Thus, even though the Adam~\cite{Adam} optimize algorithm we used can adjust the learning rate, we found a polynomial learning rate decay strategy is still useful. The hyper-parameters of it are: initial learning rate=$1\text{e-}4$, overall decay steps=20000, end learning rate=$5\text{e-}6$, power=0.5. These results are shown in Fig.~\ref{fig:s.a.4.2.c}(b).

\subsection{State and action space} \label{section: state and action}
The state space for RG and TG are show in Table~\ref{tab:2} and Table~\ref{tab:3}. The added state vector features in XfrNet are: \textit{difficulty, game\_loop, food\_army, food\_workers}. The macro-actions of the TG and RG are shown in Table~\ref{tab:action}. The added macro-actions in RG are: \textit{attack\_queued, retreat\_queued, gas\_worker\_only, attack\_main\_base, attack\_sub\_base}.

\begin{table}[ht]
    % \scriptsize
    % \small
    \centering
    \caption{Real game state features}
    \scalebox{0.95}{
    \begin{tabular}{ l | l }
        \hline
        features   &  remarks  \\
        \hline
        opponent.difficulty  & from 1 to 10 \\
        observation.game-loop & time in frames  \\
        observation.player-common.minerals & minerals  \\
        observation.player-common.vespene & gas \\
        observation.score.score-details.spent-minerals & mineral cost \\
        observation.score.score-details.spent-vespene  & gas cost \\
        player-common.food-cap & max population \\
        player-common.food-used & used population \\
        player-common.food-army  & num of army \\
        player-common.food-workers  & num of workers \\
        player-common.army-count & num of army \\
        food-army / food-workers & rate \\ 
        \hline
        * num of probe, zealot, stalker, etc & multi-features \\
        * num of pylon, assimilator, gateway, etc &  multi-features \\
        \hline
        * cost of pylon, assimilator, gateway, etc &  multi-features \\
        * num of probe for mineral and the ideal num & multi-features \\
        * num of probe for gas and the ideal num & multi-features \\
        * num of the training probe, zealot, stalker &  multi-features \\
    \end{tabular}
    \label{tab:2}}
    % \vspace{-10pt}
\end{table}

\begin{table}[ht]
    % \scriptsize
    % \small
    \centering
    \caption{Thought-game state features}
    \scalebox{0.95}{
    \begin{tabular}{ l | l }
        \hline
        features   &  remarks  \\
        \hline
        time-seconds &  always be set to 0 \\
        minerals & minerals  \\
        vespene & gas \\
        spent-minerals & mineral cost \\
        spent-vespene  & gas cost \\
        collected-mineral & mineral + mineral cost \\
        collected-gas & gas + gas cost \\
        player-common.food-cap & max population \\
        player-common.food-used & used population \\
        player-common.army-count & counts of army \\
        \hline
        * num of probe, zealot, etc & multi-features \\
        * num of buildings &  multi-features \\
        * num of gas and mineral workers & multi-features \\
    \end{tabular}
    \label{tab:3}}
    % \vspace{-10pt}
\end{table}

\begin{table}[ht]
    % \scriptsize
    % \small
    \centering
    \caption{Macro-actions}
    \scalebox{0.95}{
    \begin{tabular}{ l | l }
        \hline
        macro-actions   &  consist of actions  \\
        \hline
        Build-probe &  select base $\rightarrow$ produce probe \\
        Build-zealot &  select gateway $\rightarrow$ produce zealot\\
        Build-Stalker &  select gateway $\rightarrow$ produce stalker\\ 
        Build-pylon  &  select probe $\rightarrow$ build pylon \\
        Build-gateway & select probe $\rightarrow$ build gateway \\
        Build-Assimilator  & select probe $\rightarrow$ build assimilator \\ 
        Build-CyberneticsCore & select probe $\rightarrow$ build cybernetics \\
        Attack &  select army $\rightarrow$ attack enemy base\\
        Retreat &  select army $\rightarrow$ move our base \\
		Do-nothing & None \\
    \end{tabular}
    \label{tab:action}}
    % \vspace{-10pt}
\end{table}

\section{Discussion}
\subsection{Tradeoff for building a TG}
Our method builds a hand-designed model, the way of which is somewhat considered time-consuming and burdened. We think there is a tradeoff of using it, especially in a complex environment. Previous fastest works use 1 day for training. However, this is not the full time they cost for the experiments. In machine learning, researchers always need dozens of times to tune for the best hyper-parameters~\cite{DeepLearning2015Nature}. For a complicated game, settings other than hyper-parameter need be selected carefully~\cite{vinyals2017sc2}, e.g, the actions assemble in macro-actions should whether be queued? The third phenomenon is that RL algorithms tend to have high variance~\cite{DRLmatters2018} and are advised to run for above 3 times. All these cases make the overall training time to roughly be $24 \cdot N$ (in hours), where $N$ is often larger than 10 (in our experiments, $N$ is larger than 100). We use about $72$ hours to build the TG and the training costs $1$ hour. Therefore, the following inequality $(72 + 1 \cdot N) < 24 \cdot N$ holds when $N > 3$. When the number of experiments grows, the advantage of small training time grows, mitigating the cons of preparation time. Hence, we think the tradeoff for building a TG is acceptable in the case of training an RL agent in a complex environment.

\subsection{TG vs. Imitation learning}
One advantage of TG is to prepare a good initial policy for real tasks. IL can also provide an initial policy. The difference is that: IL gives an answer, and the agent is trained by imitating it, while TG gives a question, making the agent trained through solving the simpler question. In the real world, humans can gain advantages from both ways. Even though the mainstream at now is using IL to get a good initial policy, we advocate that not to overlook the methods of proposing simpler questions. We argue that TG can have a role in such situations: 1. No human expert trajectories exist, or the cost to collect them is high; 2. The speed of the original simulator is slow, leading to a long time of training; 3. The original reward is sparse, making learning difficult.

\section{Conclusion}
This paper discloses an interesting phenomenon that TG is a useful way to inject human knowledge into RL. We use extensive experiments to show the effectiveness of TG on the SC games and another real-world task. We come up with the XfrNet which can handle the cases when needing to add observations and actions in TL. XfrNet is not only suitable for TG, but also in the TL tasks when states and actions are different. We introduce a TG hypothesis and use experiments in SC2 and another task to validate it. This paper also concludes the tricks and experiences in training on SC which can not only give insight for SC but also bring inspiration for other complex scenarios.

% or


\section*{Acknowledgment}
This work was supported by the Natural Science Foundation of China under Grant 61672273 and Grant 61832008.

\ifCLASSOPTIONcaptionsoff
  \newpage
\fi



%\bibliographystyle{IEEEtran}
\bibliographystyle{IEEEtran}
\bibliography{tg_1}



% if you will not have a photo at all


%\vfill


\end{document}